\pdfoutput=1

\documentclass[11pt]{article}

\usepackage[final]{acl}
\usepackage{booktabs}
\usepackage{times}
\usepackage{latexsym}
\usepackage{url}
\usepackage{enumitem}
\usepackage{listings}
\usepackage{multirow}
\usepackage{enumitem}
\usepackage{adjustbox}
\definecolor{codegreen}{rgb}{0,0.6,0}
\definecolor{codegray}{rgb}{0.5,0.5,0.5}
\definecolor{codepurple}{rgb}{0.58,0,0.82}
\definecolor{backcolour}{rgb}{0.95,0.95,0.92}

\definecolor{maria-color}{HTML}{7881F2}

\newcommand{\rev}[1]{\textcolor{black}{#1}}

\usepackage{multirow}
\usepackage{graphicx,subcaption}

\usepackage{pgf}
\usepackage{colortbl}

\definecolor{color1}{HTML}{93003a}
\definecolor{color2}{HTML}{cf3759}
\definecolor{color3}{HTML}{f4777f}
\definecolor{color4}{HTML}{ffbcaf}
\definecolor{color5}{HTML}{ffffe0}
\definecolor{color6}{HTML}{a5d5d8}
\definecolor{color7}{HTML}{73a2c6}
\definecolor{color8}{HTML}{4771b2}
\definecolor{color9}{HTML}{00429d}

\usepackage{array}

\newcommand*{\opacity}{30}%
\newcommand*{\minval}{-15}%
\newcommand*{\maxval}{15}%

\newcommand{\gradient}[1]{
    \ifdimcomp{#1pt}{>}{\maxval pt}{#1}{
        \ifdimcomp{#1pt}{<}{\minval pt}{#1}{
            \pgfmathparse{int(round(8*(#1/(\maxval-\minval))-(\minval*(8/(\maxval-\minval)))))}
            \xdef\tempa{\pgfmathresult}
            \ifcase\tempa
                \cellcolor{color1!\opacity} #1\or
                \cellcolor{color2!\opacity} #1\or
                \cellcolor{color3!\opacity} #1\or
                \cellcolor{color4!\opacity} #1\or
                \cellcolor{color5!\opacity} #1\or
                \cellcolor{color6!\opacity} #1\or
                \cellcolor{color7!\opacity} #1\or
                \cellcolor{color8!\opacity} #1\or
                \cellcolor{color9!\opacity} #1
            \fi
    }}
}

\newcommand{\gradientspecificity}[1]{
    \ifnum#1>0
        \cellcolor{green!10} #1
    \else
        \cellcolor{red!10} #1
    \fi
}

\usepackage{lipsum}
\usepackage[T1]{fontenc}

\usepackage[utf8]{inputenc}

\usepackage{microtype}

\usepackage{inconsolata}

\usepackage{graphicx}

\graphicspath{{../images/}}

\title{Multi-Modal Framing Analysis of News}

\author{Arnav Arora\thanks{~~denotes equal contribution}\hspace{0.5cm} Srishti Yadav\footnotemark[1]\hspace{0.5cm} \\
\textbf{Maria Antoniak}\hspace{0.5cm} \textbf{Serge Belongie}\hspace{0.5cm} \textbf{Isabelle Augenstein} \\
 University of Copenhagen\\
  \texttt{\{aar, srya, maan, s.belongie, augenstein\}@di.ku.dk}\\
}

\begin{document}
\maketitle

\begin{abstract}
Automated frame analysis of political communication is a popular task in computational social science that is used to study how authors select aspects of a topic to \textit{frame} its reception. So far, such studies have been narrow, in that they use a fixed set of pre-defined frames and focus only on the text, ignoring the visual contexts in which those texts appear. Especially for framing in the news, this leaves out valuable information about editorial choices, which include not just the written article but also accompanying photographs. To overcome such limitations, we present a method for conducting multi-modal, multi-label framing analysis at scale using large (vision-) language models. Grounding our work in framing theory, we extract latent meaning embedded in images used to convey a certain point and contrast that to the text by comparing the respective frames used. We also identify highly partisan framing of topics with issue-specific frame analysis found in prior qualitative work. We demonstrate a method for doing scalable integrative framing analysis of both text and image in news, providing a more complete picture for understanding media bias.
\end{abstract}
 
\section{Introduction}

Frames are conceptual tools that both communicators and audiences use to interpret and categorize issues and events \cite{gitlin1980whole, eko1999framing, pan1993framing, reese2001prologue}. 
By highlighting specific elements of a topic and minimizing others, communicators \textit{frame} messages in ways they believe will resonate with audiences \cite{goffman1974frame} and can shape the way the topic is perceived by readers or viewers~\citep{schudson2003sociology}. In the field of journalism, framing is a core narrative device by which news consumption is framed within an interpretive perspective~\citep{card2015media}. Most prior work on computational frame analysis has focused on linguistic structure and content analyses via text elements~\citep{ali_survey_2022}.
However, framing is not solely textual; visual elements also play a crucial role in conveying implicit and explicit messages~\citep{messaris2001role}.

\begin{figure}[t]
    \centering
    \includegraphics[width=\linewidth]{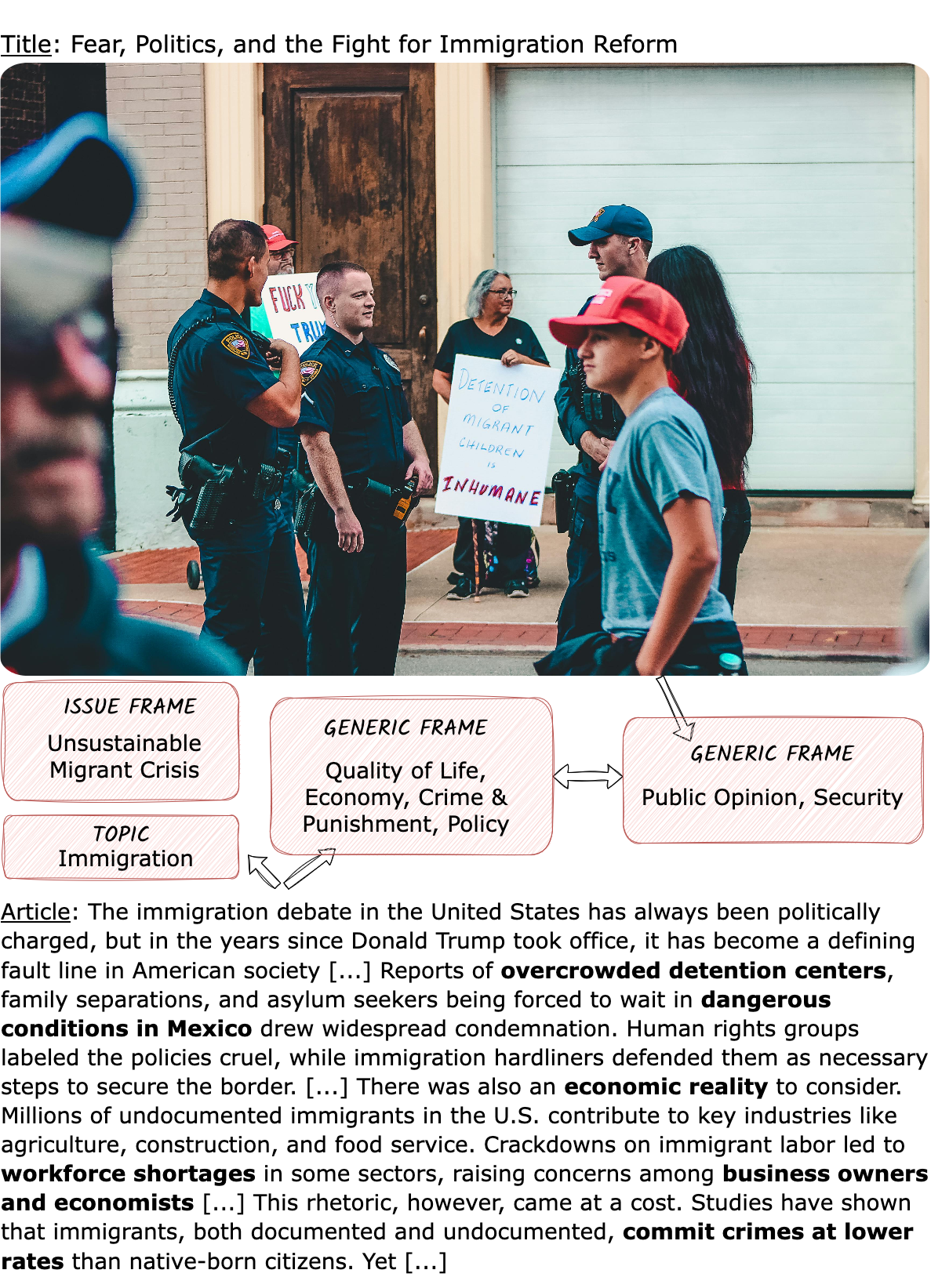}
    \caption{\rev{News can be intentionally \textit{framed} to affect reader perception. Editorial choices decide what is communicated through words and images. Our approach systematically detects this framing.}}
    \label{fig:migrant-crisis}
\end{figure}

News media often employ images alongside text to reinforce or contrast the intended frame, leveraging the affective and cognitive impact of visuals \cite{cope2005image,keib2018picture,grabe2009image,geise2025news}. An example can be seen in \autoref{fig:migrant-crisis}. The image depicts protesters with signs with police presence around, using the public opinion and security framing. The article text, on the other hand, talks about the quality of life of migrants, economic implications, crime, and policy framing. Such differences in portrayal can affect readers' perception of the crisis and their induced emotions. Further, while communication in text is more explicit through linguistic framing, images encode frames in more subtle and implicit ways, requiring sophisticated interpretation models to capture their meaning \cite{aiello2019visual}. As such, when conducting media analysis, the communicated framing across both the image and the text should be considered.

While visual framing analysis has been explored in communication science and journalism studies~\citep{wessler2016global,powell2017multimodal}, existing studies doing computational framing analysis have largely ignored this crucial aspect~\citep{ali_survey_2022}. Further, they have primarily focused on a fixed set of labels for framing analysis, performing prediction in a multi-class setting. This substantially limits the information one can derive from predictions as an article can convey several frames (\autoref{fig:migrant-crisis}) and fine-grained analysis of framing within a topic necessitates frames specific to a particular issue. Large vision and language models are particularly well suited for conducting this task at scale, considering the semantic understanding embedded into them through large scale pre-training. \rev{This led us to our overall research questions: \textbf{(RQ1)} Can LLMs and VLMs reliably detect framing in news articles? \textbf{(RQ2)} Are there differences in framing conveyed through text vs the images? and \textbf{(RQ3)} How do these framings vary across topics and publishers?}

Our \textbf{contributions} include the following.
 
\vspace{-0.5em}
\begin{itemize}[noitemsep,leftmargin=*]
\item We present the first computational study of multimodal framing in the news, outlining a methodology using LLMs and VLMs.
\item We provide a large-scale dataset of 500k U.S.-based news articles for framing analysis, automatically labeled with generic and issue-specific frames, validated through human annotations.\footnote{\url{https://huggingface.co/datasets/copenlu/mm-framing}}
\item We conduct a thorough analysis of frames communicated in the image vs the text of articles, both at a corpus-level and at a fine-grained level for articles about immigration. 
\item We discover meaningful differences in how frames are used in images and texts, across political leanings and topics.
\end{itemize}
\vspace{-0.5em}

\section{News Framing}

Framing of news articles has been studied widely in communication studies. There are several definitions of framing and scholars often disagree on the method for extraction and its role in public communication, leading to it being termed a ``fractured paradigm''~\citep{entman1993framing}. The definition from Entman for framing at large, however, is the most widely accepted one. 
He defined framing as \textit{``making some aspects of reality more salient in a text in order to promote a particular problem definition, causal interpretation, moral evaluation, and/or treatment recommendation for the item described''}. Narrowing the framework to news, ~\citet{de2005news} lays out a typology of news framing. He states that there are \textit{generic} news frames and \textit{issue-specific} news frames. \textit{Generic} news frames \textit{``transcend thematic limitations and can be identified in relation to different topics, some even over time and in different cultural contexts''}. They are particularly useful for uncovering broad patterns within or across countries. Issue-specific frames allow for richer, more fine-grained analysis of various aspects highlighted within a particular issue.

\paragraph{\rev{Framing Through Visuals}}

Photographs are an important vehicle of framing as a reader may process textual and visual messages differently. Readers may focus on photographs without also reading an accompanying story \cite{miller1975content} or might select which news story to read depending on the image thumbnail. Images are potent framing tools because they evoke immediate emotional responses, provide contextual cues, and sometimes contradict textual narratives \cite{geise2024effects}. For visual data, however, automation is more challenging because computational image analysis often struggles with connotative and symbolic elements that are readily discernible to human annotators~\cite{rodriguez2011levels}. While some recent advancements in machine learning and computer vision offer promising avenues for automated visual framing analysis, the complexity of symbolic and ideological meanings typically requires human interpretation \cite{coleman2010framing}.

\paragraph{Automated Framing Analysis}

In computational studies, framing has been studied with the help of machine learning methods for content analysis at scale. ~\citet{ali_survey_2022,vallejo_connecting_2023,otmakhova-etal-2024-media} survey these efforts towards on computational framing analysis, focusing on different aspects like methods, varying conceptualisations and inter-disciplinary connections of framings. 
The most widely used resource is one by ~\citet{card2015media}, who present the Media Frames Corpus (MFC), a dataset of US based news headlines annotated for generic frames. Most studies use it in a supervised, multi-class prediction setting with MFC frames for analysis of social networks~\citep{mendelsohn_modeling_2021} or discussion forums~\citep{hartmann_issue_2019}. There are also English-only and multi-lingual SemEval tasks on frame detection~\citep{piskorski_semeval-2023_2023, piskorski_multilingual_2023, sajwani_frappe_2024}.
Unsupervised approaches to framing analysis use framing lexica~\citep{field_framing_2018}, clustering~\citep{doi:10.1177/0894439315596385, ajjour_modeling_2019}, and topic models~\citep{nguyen_guided_2015}.
All of these approaches have focused on texts alone. 

\paragraph{Integrative Framing Analysis}

Integrative framing analysis is when both images and text are observed separately, but the results are integrated to form a more complete picture of framing analysis~\citep{danIntegrativeFramingAnalysis2017}. Although there is broad agreement that visual and verbal elements should be studied side by side \citep{coleman2010framing, rose2022visual}, relatively few studies have effectively combined the two. 
Understanding multimodal framing is important for several reasons: it allows for a more granular examination of media bias, as textual analysis alone may miss the ideological and emotional undertones of visual elements \cite{wessler2016global,geise2025news}. It improves fact-checking and detection of misleading content by identifying inconsistencies between textual claims and their supporting images. It also finds applications in political communication, journalism studies, and public policy through tracing changes in framing over time and across outlets \cite{baumgartner2008decline}.

\begin{figure}
    \centering
    \includegraphics[width=\linewidth]{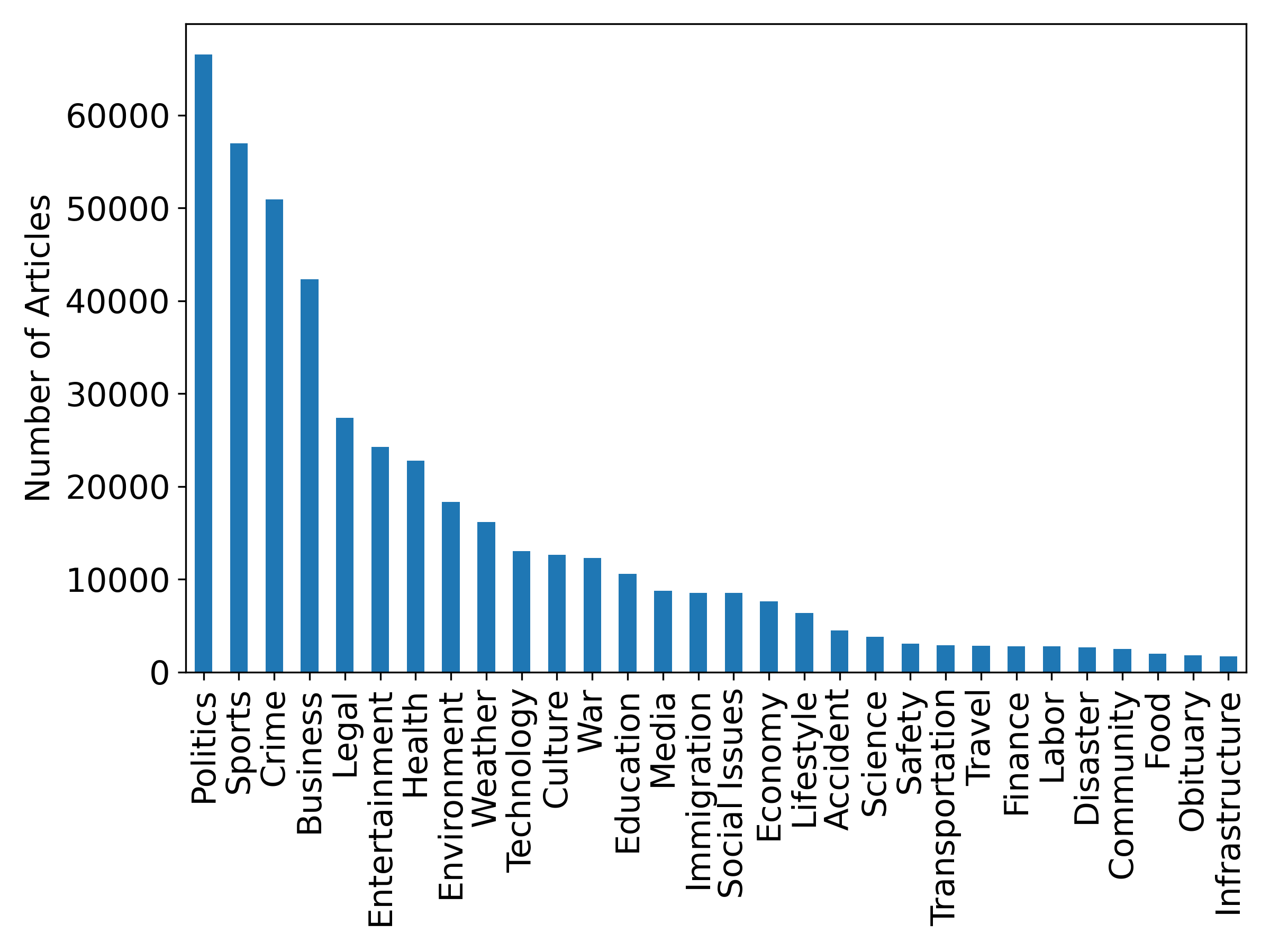}
    \caption{Distribution of data across the top 30 topics}
    \label{fig:data-dist-topic}
\end{figure}

\section{Dataset}
\label{section:dataset}

\begin{figure}
    \centering
    \includegraphics[width=\linewidth]{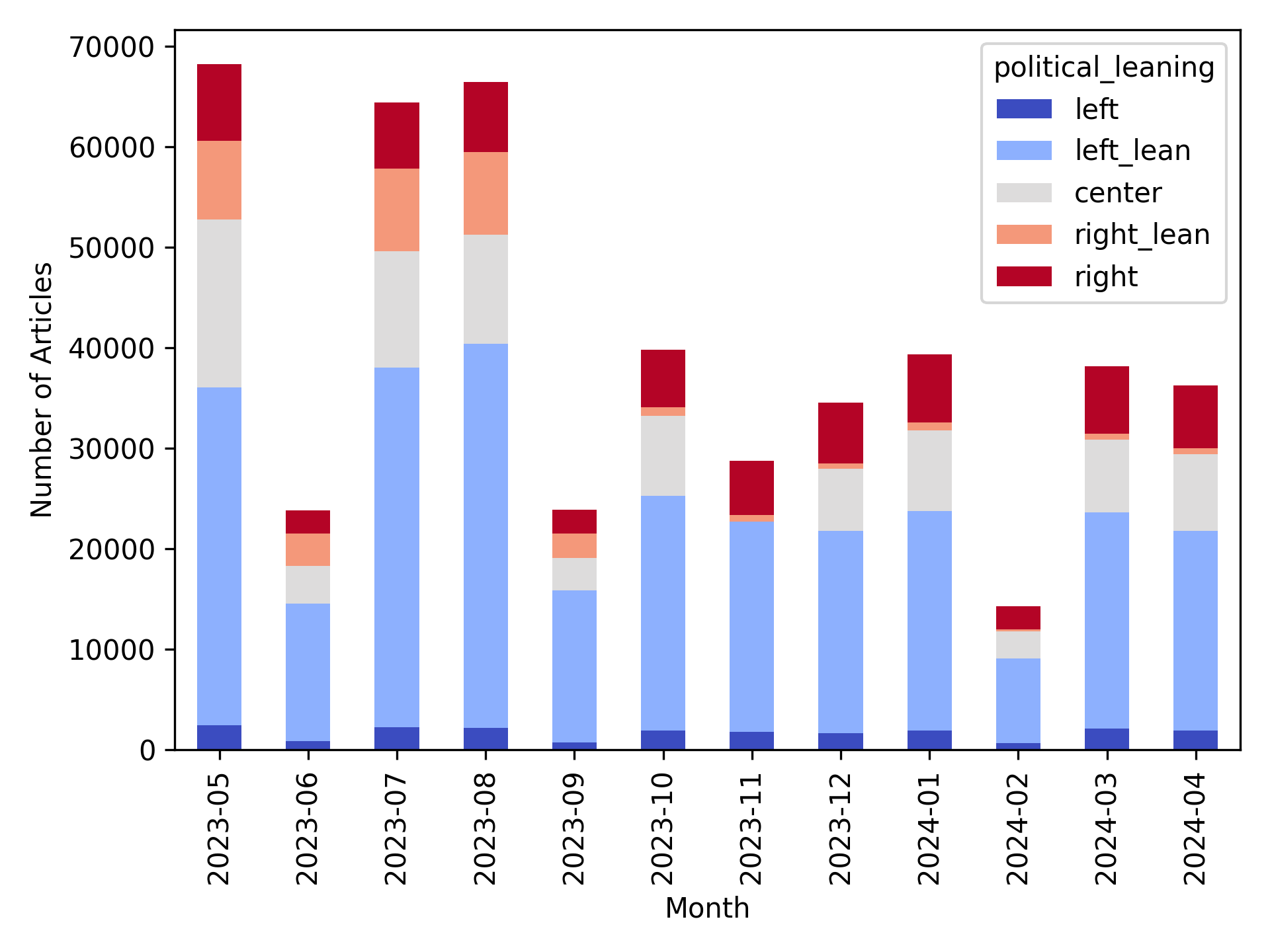}
    \caption{Distribution of data across the time-period of collection, broken down by political leaning.}
    \label{fig:data-dist-month}
\end{figure}

We crawl news articles along with corresponding images from 28 US-based news agencies, extracting data for a 12 month period between May 2023 and April 2024 using the \texttt{newsplease} library~\citep{Hamborg2017}. 
Our selected news sources reflect the entire political spectrum, based on data from AllSides Media Bias,\footnote{\url{https://www.allsides.com/media-bias/ratings}} an organization that assigns a rating of political leaning to each media outlet. 
Our list of sources along with corresponding political leanings are listed in~\autoref{sec:source_domains}. 
We query the publicly available Common Crawl archives\footnote{\url{https://commoncrawl.org/news-crawl}} for the corresponding publishers and extracted each article’s text, headline, publication date, image\_urls and other metadata in JSON format. 
Post scraping, we filter extremely short and long articles, images of logos and other noise. Further details about the filtering process are provided in \autoref{app:filtering}. 
Our final dataset includes about 500K articles and corresponding images.
We show the distribution of the data across the time period of data collection in \autoref{fig:data-dist-month}. 

\section{Method}

\subsection{Model Annotation}
\label{subsec:model_anno}
We use Large Language Models (LLM) and Vision-Language Models (VLM) to label several aspects of the text and image from news articles, including generic, issue-specific framing and the topic of the article. In \autoref{app:extracted_aspects}, we list the different aspects extracted per modality from the news articles. We also generate explanations for the decisions along each prediction for reducing hallucination and facilitating downstream qualitative analysis. For extracting generic frame across both modalities, we use the framing dimensions outlined by~\citet{Boydstun2014TrackingTD}, as provided in the Media Frames Corpus~\citep{card2015media}, which includes 15 generic frames appropriate for analysis of US news. These frames are \textit{Economic, Capacity \& Resources, Morality. Fairness \& Equality, Legality, Constitutionality \& Jurisprudence, Policy Prescription \& Evaluation, Crime \& Punishment, Security \& Defense, Health \& Safety, Quality of Life, Cultural Identity, Public Opinion, Political, External Regulation \& Reputation}
and \textit{Other}. The descriptions of each of the frames are provided in \autoref{app:frame_desc}. Differing from most previous studies on automated framing analysis, which assign a single frame to an article, we conduct the frame extraction in a multi-label setting, i.e., an article or an image can have more than one frame. This setting is much closer to the setup scholars use when conducting qualitative studies on framing~\citep{danIntegrativeFramingAnalysis2017}.

\subsection{Models and Prompts}
We use Mistral-7B for text annotations and Pixtral-12B (multi-modal) for image. We use the vLLM library for high-throughput inference, allowing us to conduct analysis on the entire dataset in a few days, demonstrating the scalability of our approach. See \autoref{app:model-exp-details} for more details.

We carefully craft prompts for the extraction of each aspect from the article. For extraction of frames expressed in the articles and images, \rev{we experiment with including the Entman and Gamson definitions of framing. We also experiment with short, medium, and long descriptions of each frame and different output formats. We benchmark these different prompts on a validation set and qualitatively examine model decisions and explanations, iteratively improving the prompt by based on categories of errors.} For instance, the Pixtral model has a tendency to predict the \textit{economic} frame every time an entity in professional attire appears in the image, associating professional attire with being wealthy -- we instruct the model to avoid such reliance. We report performance on a held-out test set in \autoref{subsec:model_eval}.
\rev{For issue-specific frames, we prompt the model by providing a definition of framing and some examples of issue-specific frames across topics. The task was then to analyse the article and generate an issue-specific frame (described in 2-3 words) w.r.t. the topic of the article. More details are provided in \autoref{app:issue_frame_analysis}.}
The full list of prompts for the extraction of each aspect is provided in Appendix ~\autoref{lst:prompt-image-frame-prediction} and ~\autoref{lst:text-frame-prediction}.

\begin{figure}[]
    \centering
    \includegraphics[width=\linewidth]{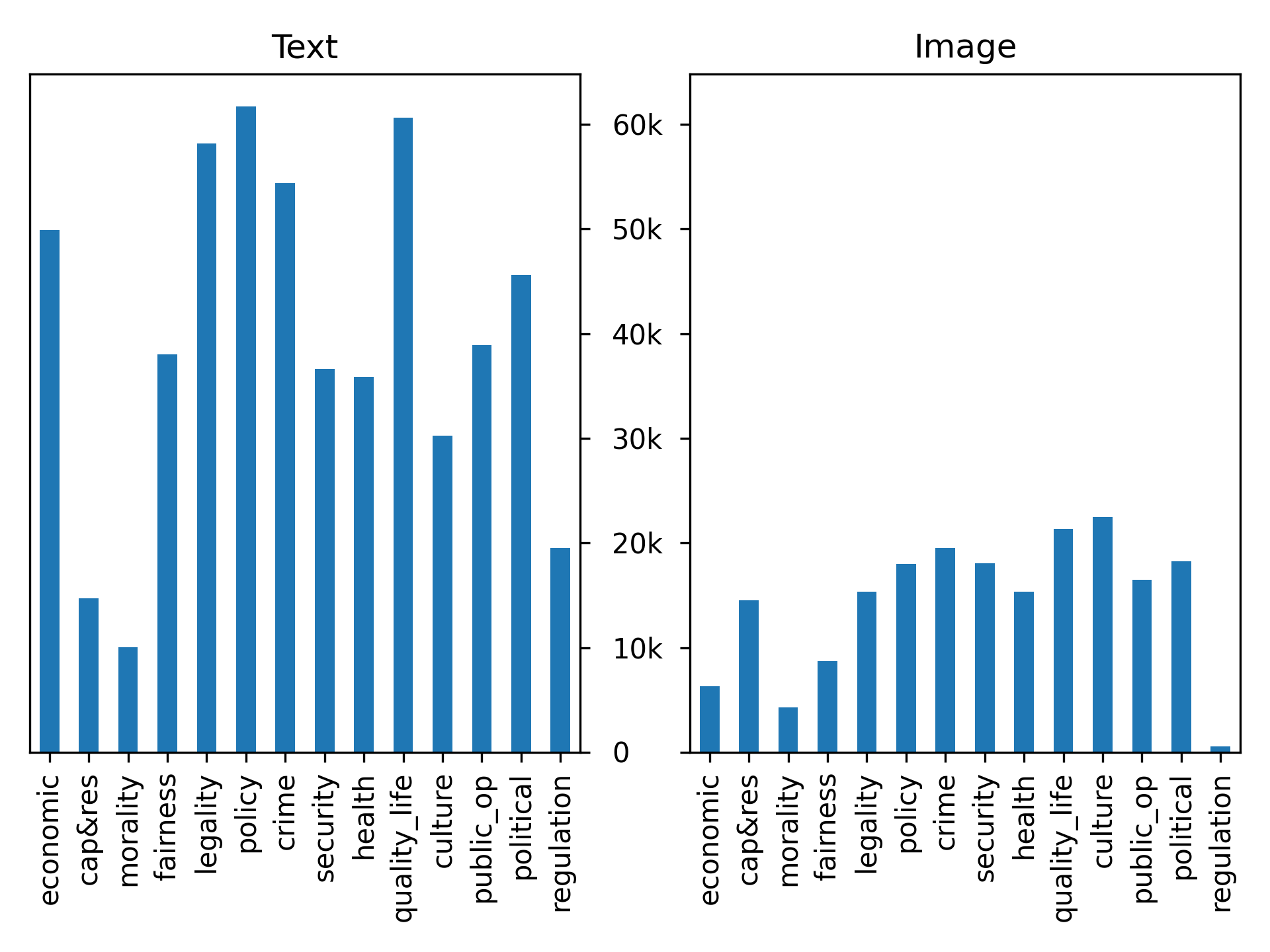}
    \caption{\rev{Frequency of predicted generic frames across all articles for texts and images.}}
    \label{fig:frame-freq-all}
\end{figure}

\subsection{Model Evaluation}
\label{subsec:model_eval}

\paragraph{Text Framing} To assess the quality of output frames from our approach, and compare against prior work, we evaluate our text frame extraction model on the existing large scale benchmark Media Frames Corpus~\citep{card2015media}. 
This corpus provides frame labels by each annotator for over 32,000 articles from US news on topics like immigration, smoking, and same-sex marriage. We take the union of all frame labels assigned by the annotators and assign the top 3 most frequent frames assigned by the annotators as the labels for an article. We run our text framing analysis model on this dataset, providing the article of the text while allowing the model to generate multiple frames per article. We calculate the intersection between the sets of Mistral model annotated frames and human-annotated frames for each article in the dataset. 95.7\% of the articles had a non-zero intersection, with at least one overlapping frame label, demonstrating that the model outputs accurate frames in most cases. Our model received a micro averaged F1 score of 0.5, with an averaged precision of 0.42 and averaged recall of 0.62. For prediction across 15 labels, in a multi-label setting, and a task as subjective as framing, we believe the model performs quite well, \rev{with several errors being attributed to subjective interpretation. We provide an error analysis of the model with per-label metrics, frequent misclassifications and their examples in \autoref{app:frame-class-perf}.}

\paragraph{Image Framing}

For images, there is no existing benchmark with frame labels. 
Two of the authors of this study manually annotated 600 images for generic frames across a stratified (time-period and topics) sample of the dataset.
We use annotation guidelines released by~\citet{mendelsohn_modeling_2021}, adapting them to the image annotation setting with examples of images for each framing category. We set up a multi-label classification platform (\autoref{fig:image-annotation-ui}) where annotators select one or more frames from the list given an image.
 \rev{To calculate agreement, we compute both Krippendorff’s alpha ($\alpha = 0.393$) and mean Jaccard Index (0.614) as measures of inter-annotator agreement. While Krippendorff’s alpha provides label-level reliability, Jaccard Index is particularly well-suited to our multi-label setting, as it evaluates instance-level agreement and gives partial credit when annotators agree on some but not all labels.} Framing analysis is a highly subjective task, as is well established by prior work in communication studies as well as NLP~\citep{card2015media}. For images, the subjectivity increases substantially, given the limited amount of context available and requiring cultural and conceptual familiarity more so than needed for textual framing analysis~\citep{geise2015-link}. Our agreement scores are thus in line with prior work on framing.

\rev{To further minimise the effect of subjectivity of the annotations on our findings, we take the union of the frame annotations by two annotators as our gold set. The model is thus tasked with generating \textit{all} labels annotated by the different annotators. We calculate the intersection of model predictions and the gold set for each image. The proportion of non-zero intersection instances between the model and human annotations is 84.2\%, i.e., at least one correctly predicted frame most of the time, demonstrating the model aligning with the human framing interpretations in a majority of the cases, with the most frequent error being a `None' prediction. We conduct a thorough analysis of frequent mis-classifications by the model in \autoref{app:frame-class-perf}.}

\begin{figure*}[ht!]
    \centering
    \includegraphics[width=\linewidth]{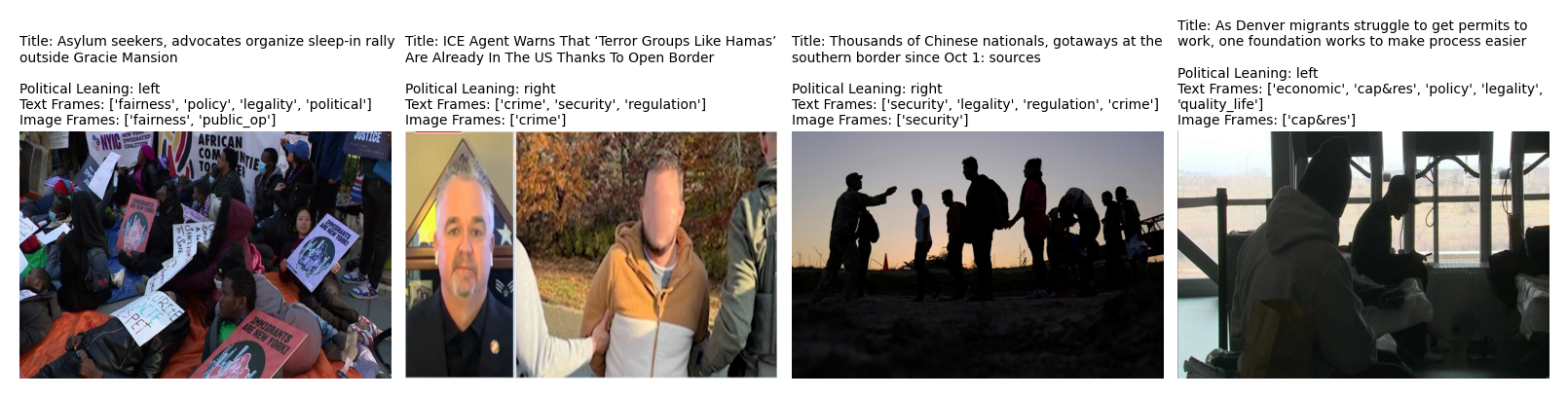}
    \caption{
    Examples of generic frame prediction in images vs texts about immigration across political leanings.
    }
    \label{fig:immigration-frame-images}
\end{figure*}

\paragraph{Topics}
To evaluate the topics output by the model, we use the 20 most frequent topics for our analysis, discarding one (\textit{Media}) because most of its included documents were publication boilerplate.
Two of the authors hand-annotate a set of 190 articles (10 per topic), marking whether the model's topic prediction was acceptable. \rev{The overlap between the annotators was 83.5\%, substantially higher than random chance (50\%).}
When calculated against the labels assigned by each annotator, the overall accuracy for topics predicted by the model was found to be 86\% and 87\%.
We list these, with examples and accuracies, in \autoref{table:topics}.

\begin{figure*}[h]
    \centering
    \includegraphics[width=\textwidth]{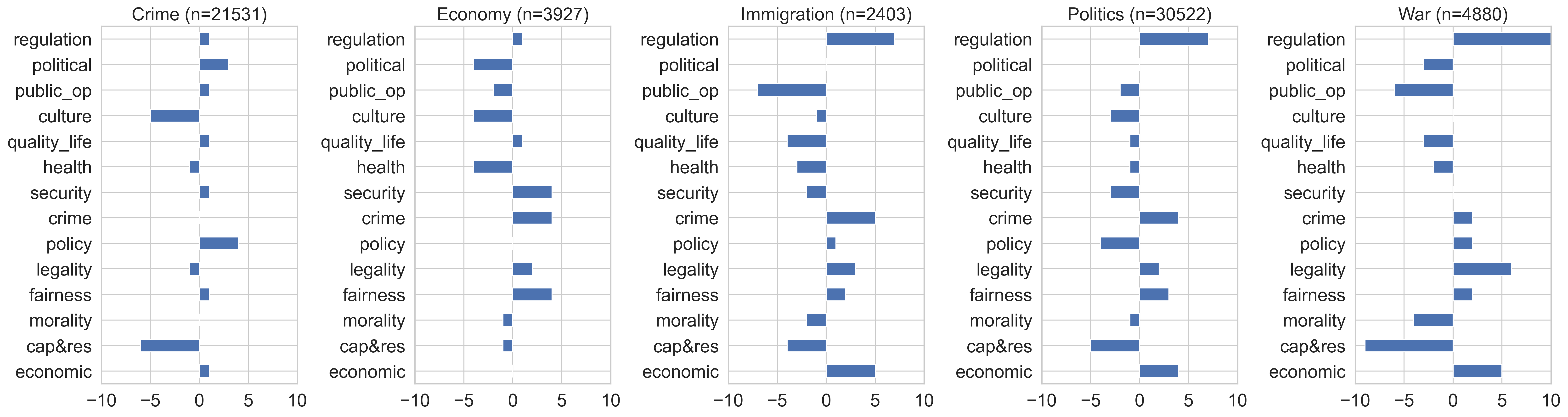}
    \caption{
    Comparison of generic frame prediction frequencies in images vs texts. The x-axis represents the subtracted rank of predictions for that frame between the two modalities: \textbf{positive} scores indicate that the frame was more often predicted in \textbf{texts}, \textbf{negative} scores indicate that the frame was more often predicted in \textbf{images}.
    }
    \label{fig:frame-rank-sel-topics}
\end{figure*}

\section{Multi-Modal Framing Analysis}

\subsection{Generic Framing Analysis}

We first analyze the generic frames predictions across the dataset\footnote{For framing analysis, we perform additional filtering,  excluding articles with `None' frame predictions. We further removed articles less than 100 words and articles on the topics `sports' and `media', as we observed while conducting qualitative analysis that articles on sports only used the \textit{Cultural Identity} frame in the images and several articles with the media topic only had paywall text. This resulted in a reduced set of 154k articles} in ~\autoref{fig:frame-freq-all}. Overall, we can see that there are many more predicted frames for text compared to the images (a mean of 3.6 predicted frames per text and 1.3 predicted frames per image). This is intuitive (and can be seen in the ~\autoref{fig:migrant-crisis}) since the text of the article can more easily express several distinct frames. 
Looking at their \rev{relative distributions}, we can see that in the text of the article, \textit{quality of life}, \textit{legality}, and \textit{policy} appear more frequently, while \textit{morality} and \textit{capacity \& resources} are more rare. 
In terms of image frame distribution, \textit{external regulation \& reputation} has very low frequency, while \textit{culture} and \textit{quality of life} are the most frequent. 
Contrasting the two modalities, \textit{economic} and \textit{policy} frames much more frequently appear in text, while \textit{capacity \& resources} and \textit{culture} are more prevalent in images.

We also compare frames by modality across topics, as shown in~\autoref{fig:frame-rank-sel-topics}. 
We observe substantial differences in how topics are framed across images and texts.
When covering war,
news outlets focus more on the \textit{external regulation \& reputation} and \textit{crime} framing in the text, but the images that are used convey \textit{public opinion} and \textit{capacity \& resources} framing, depicting people giving speeches and of military equipment. 
For articles on the topic of economy, there is more focus on \textit{fairness} and \textit{security \& defense} spending in texts, while the images depict the \textit{health and safety}, \textit{culture}, and \textit{political} frames. 

\begin{figure}[t]
    \centering
    \includegraphics[width=0.9\linewidth]{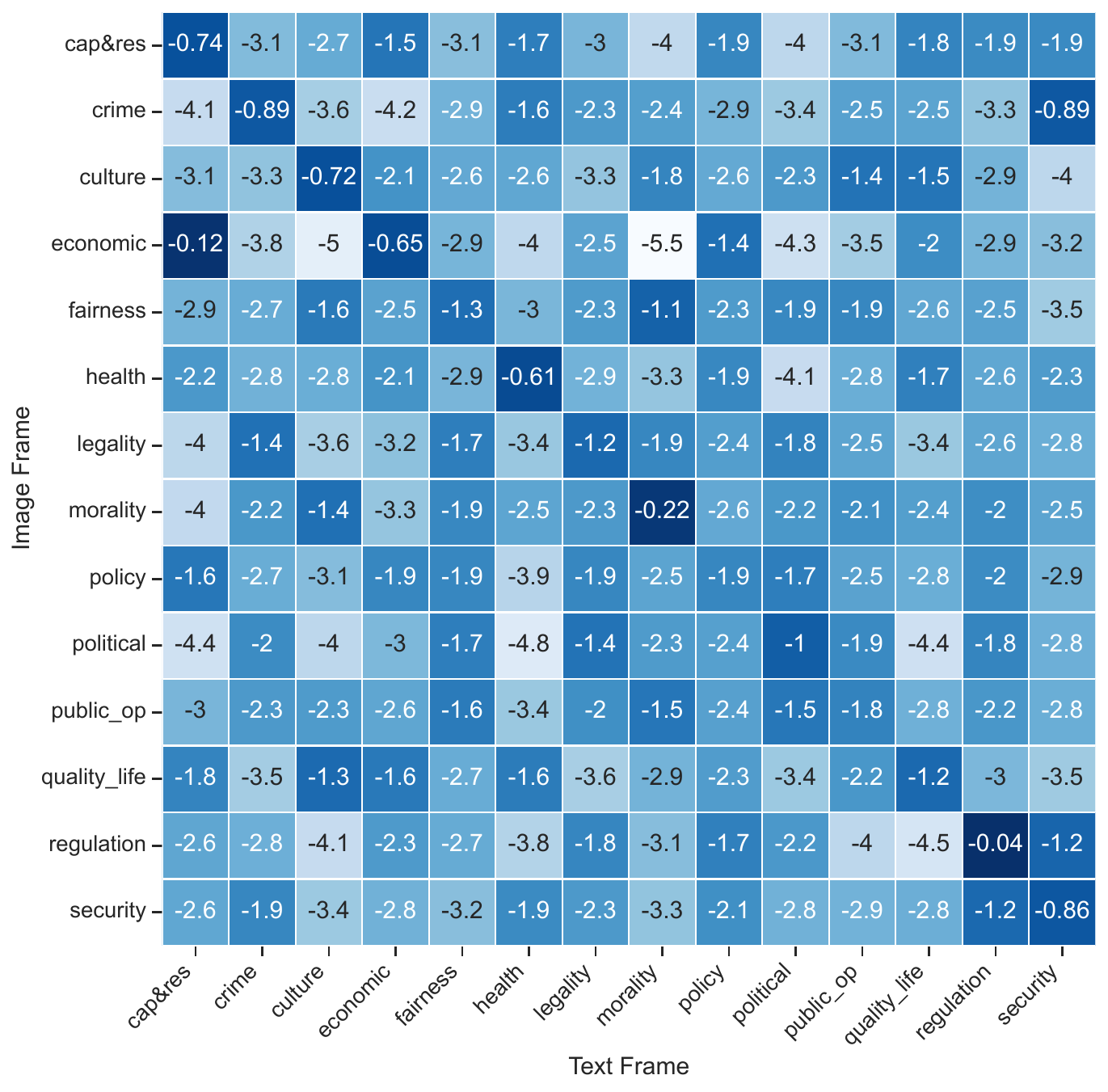}
    \caption{Pointwise mutual information between text and image frames across the dataset. Some frames are used consistently in texts and images for the same article (dark cells in the diagonal), other frames differ widely.}
    \label{fig:co-oc_matrix}
\end{figure}

\paragraph{Frame Co-Occurrence}

\begin{figure*}[h]
    \centering
    \includegraphics[width=\textwidth]{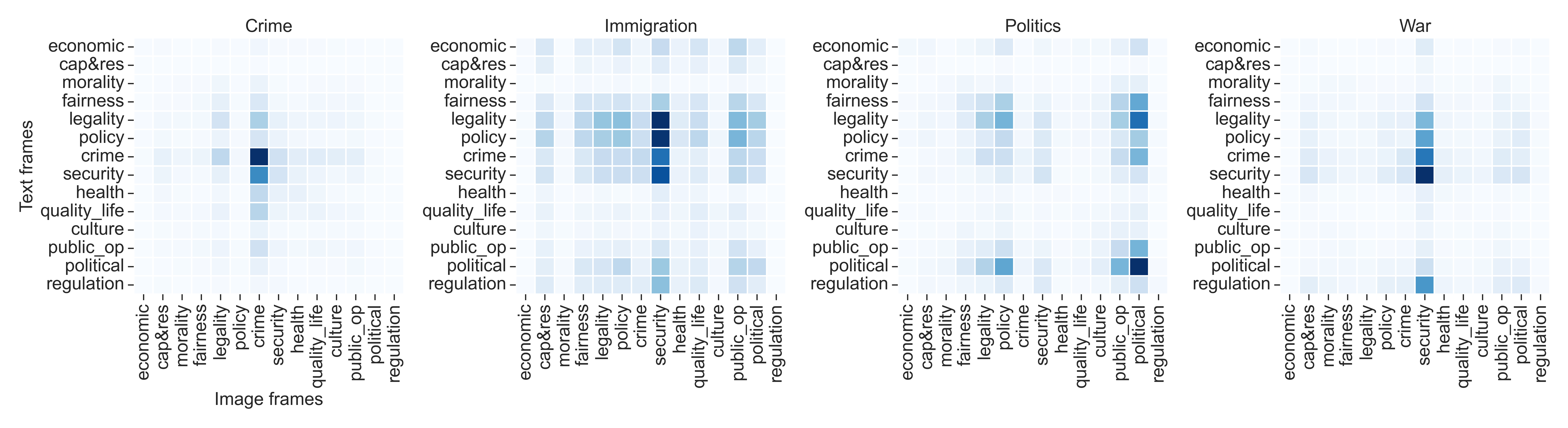}

    \caption{Frequency of frame co-occurrence between text and image frames across four selected topics.}
    \label{fig:co-oc_matrix_sel_topics}
\end{figure*}

To understand what is highlighted in the text of the article compared to the image, we plot the pointwise mutual information (PMI) of image frames and text frames across the entire dataset in~\autoref{fig:co-oc_matrix}. 
We can see the presence of a diagonal, demonstrating that there is often alignment between frames across modalities.
However, there are many deviations, some of which are intuitive. For example, depictions of \textit{quality of life} in the images when writing about cultural topics and vice versa is quite common. \textit{Quality of life} framing in the images is also used when the text is using an \textit{economic} frame. 
\textit{Political} framing in the text is associated with \textit{policy} and \textit{public opinion} framing, and \textit{legal} framing in text is associated with \textit{political} framing in the images.
To get a finer-grained understanding of co-occurrence of frames across the modalities, we analyse the percentage of co-occurrences per topic, as shown in \autoref{fig:co-oc_matrix_sel_topics}. 
For articles about crime, when using the \textit{criminal} framing in images, the texts also tend to highlight the \textit{security}, \textit{quality of life}, and the \textit{legal} frames. 
For war, the images consistently highlight the \textit{security \& defense} framing, even when the article text is highlighting the \textit{policy} or \textit{legality} frames.

\paragraph{Lexical Comparison}

\begin{table}[t]
    \centering
    \resizebox{1\linewidth}{!}{
    \scriptsize
    \begin{tabular}{clcl}
    \toprule
        \multicolumn{2}{c}{crime} & \multicolumn{2}{c}{quality of life}
        \\ 
        \cmidrule(lr){1-2}\cmidrule(lr){3-4}
        $z$-score & bigram & $z$-score & bigram 
        \\ 
        \midrule 
        \\ [-10pt]     
        
$\gradient{14.7}$ & year old & $\gradient{5.5}$ & disney world \\
$\gradient{13.6}$ & police said & $\gradient{5.0}$ & mother day \\
$\gradient{12.3}$ & police department & $\gradient{4.9}$ & ice cream \\
$\gradient{7.8}$ & county sheriff & $\gradient{4.8}$ & morning brew \\
$\gradient{7.5}$ & police officers & $\gradient{4.5}$ & prime day \\
$\gradient{7.3}$ & sheriff office & $\gradient{4.5}$ & walt disney \\
$\gradient{7.3}$ & old man & $\gradient{4.0}$ & memorial day \\
$\gradient{7.2}$ & police chief & $\gradient{4.0}$ & privacy policy \\
$\gradient{6.6}$ & police say & $\gradient{3.9}$ & black friday \\
$\gradient{6.5}$ & law enforcement & $\gradient{3.6}$ & day deals \\

\midrule

$\gradient{-4.4}$ & justice department & $\gradient{-3.1}$ & getty images \\
$\gradient{-4.4}$ & biden administration & $\gradient{-3.1}$ & former president \\
$\gradient{-4.6}$ & united states & $\gradient{-3.1}$ & tropical storm \\
$\gradient{-5.0}$ & hunter biden & $\gradient{-3.2}$ & police said \\
$\gradient{-5.3}$ & joe biden & $\gradient{-3.7}$ & taylor swift \\
$\gradient{-5.4}$ & president donald & $\gradient{-4.0}$ & health care \\
$\gradient{-5.9}$ & supreme court & $\gradient{-4.3}$ & interest rates \\
$\gradient{-6.0}$ & white house & $\gradient{-4.4}$ & interest rate \\
$\gradient{-7.5}$ & donald trump & $\gradient{-4.4}$ & social security \\
$\gradient{-7.8}$ & former president & $\gradient{-4.7}$ & student loan \\
        
        \bottomrule \\
    \end{tabular}
    }
    \caption{
        The bigrams most associated with the \textbf{images} (higher $z$-scores) and \textbf{texts} (lower $z$-scores) for the \textit{crime} and \textit{quality of life} frames. 
    }
    \label{table:fightin-words-condensed}
\end{table}

Our results above indicate that frames are used in images and texts in different ways, but \textit{how} those uses differ is unclear.
To explore this question, we perform a lexical analysis of the words used in articles whose image or text used the same frame.
We use the ``Fightin' Words'' algorithm from \citet{monroe2008fightin}, a comparison metric which takes into account disproportionate numbers of samples as well as rare words, to find the bigrams from the article texts most associated with a single frame for images vs texts.
Table \ref{table:fightin-words-condensed} shows the bigrams sorted by their $z$-scores (prior=$0.01$, frequency$\ge5$) for two frames.
Qualitatively, we observe that words associated more with image frames tend to be concrete (``ice cream'') and associated with a single meaning more easily recognized to the predicted frame (``police department'').
Prior work suggests that more tightly clustered and recognizable images are associated with more concrete topics \citep{hessel-etal-2018-quantifying}.
See the Appendix for a full list of the frames.

\begin{figure*}[ht]
    \centering
    \begin{subfigure}{\textwidth}
        \centering
        \includegraphics[width=\textwidth]{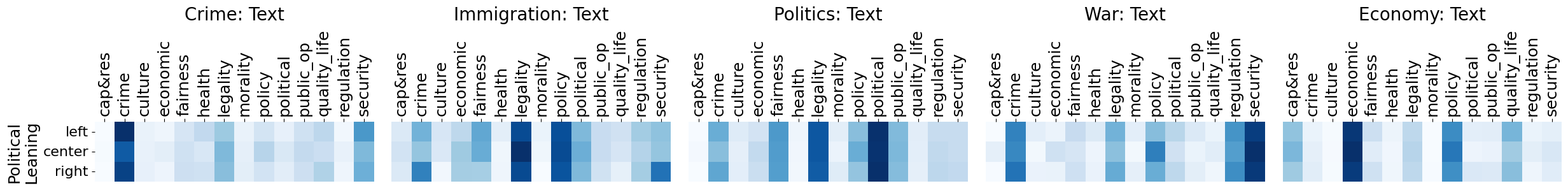}
    \end{subfigure}
    
    \vspace{0.1cm} %

    \begin{subfigure}{\textwidth}
        \centering
        \includegraphics[width=\textwidth]{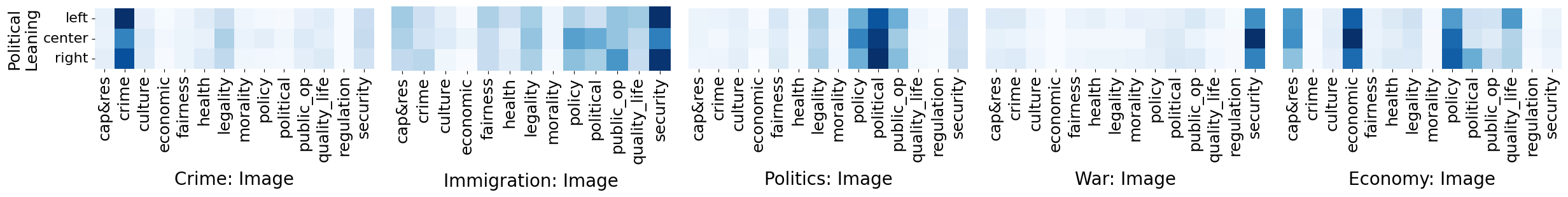}
    \end{subfigure}

    \caption{Comparison of text and image frame distributions across political leanings for five topics. 
    }
    \label{fig:political-leaning}
\end{figure*}

\paragraph{Political Leaning}
We also analyzed the correlation between political leaning and text/image frames across topics. 
For each topic, we compute how often each frame appears for each political leaning, combining \textit{left-lean} and \textit{left} as well as \textit{right-lean} and \textit{right} (see \S\ref{section:dataset}). 
\autoref{fig:political-leaning} shows the proportional frequency of frames per leaning, allowing us to compare which frames dominate among different leanings for a specific topic. 
We observe that for topics like crime and war, the images do not portray many frames other than frames like \textit{crime} and \textit{security}, unlike texts where we see additional frames like \textit{legality}, \textit{policy}, \textit{regulation} and some references to the \textit{political} frame across all political spectrum. On the other hand, topics like immigration have more variation. Right-leaning news agencies focus more on \textit{legality}, \textit{policy,} and \textit{security} frames in text and on \textit{public opinion} and \textit{security} frames in images. Similarly for politics, apart from the \textit{political} frame, images unlike texts do not often feature the \textit{crime} frame.

\subsection{Case Study: Immigration}
\label{subsec:immigration_case_study}

So far, we demonstrated the differences in generic framing across across modalities, topics, and political leaning across an entire corpus of articles. 
But our approach also allows for a more fine-grained analysis, and we demonstrate this in a case study focused on one topic: immigration.

\begin{figure*}
    \centering
    \includegraphics[width=\linewidth]{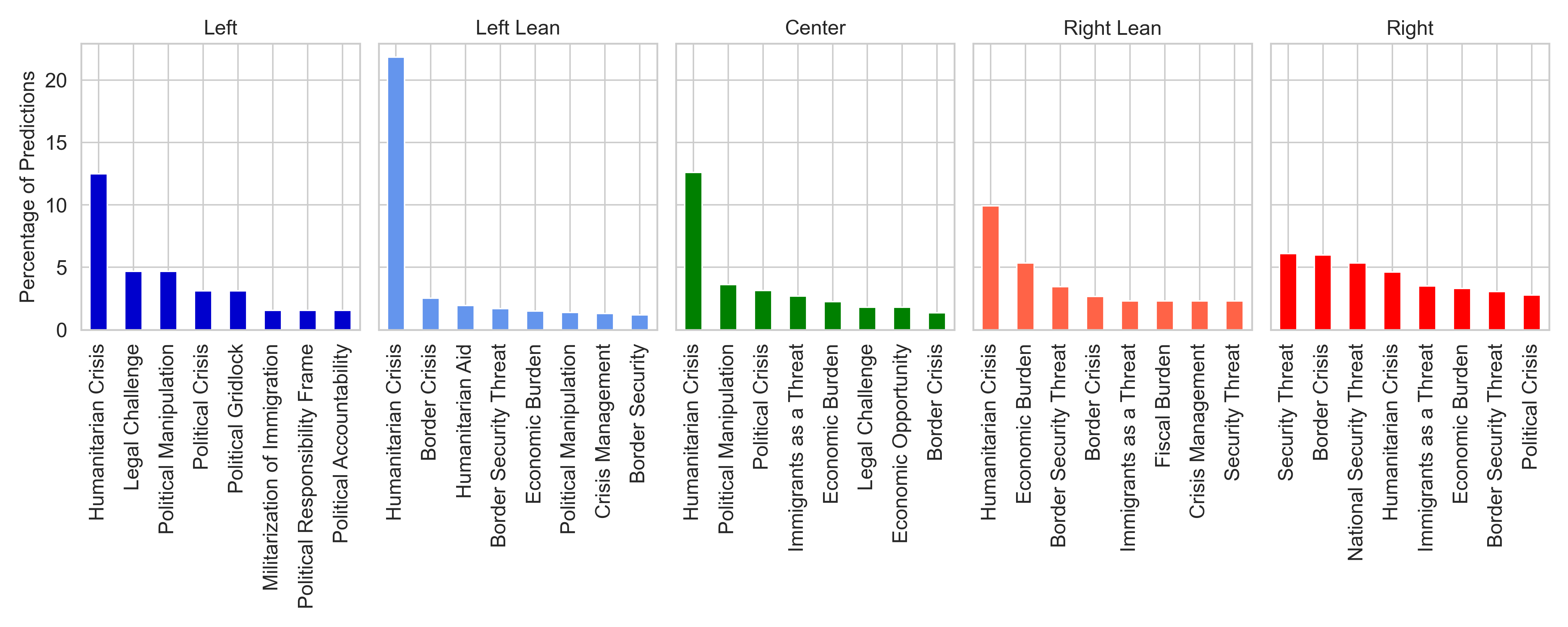}
    \caption{Normalised count of top 10 issue frames used by news publishers in the article text across the political spectrum in their reporting on immigration.}
    \label{fig:issue-frame-leaning}
\end{figure*}

In \autoref{fig:frame-rank-sel-topics}, we show that articles about immigration use the \textit{crime and punishment}, \textit{external regulation \& reputation}, and \textit{economic} framing much more frequently than the images.
These articles often focus on deals with other countries, their contribution to the economy, the cost of their deportation, and/or the crimes that they commit. 
On the other hand, images tend to use the \textit{capacity \& resources} framing, showing migrants in camps, or the \textit{public opinion} framing, showing people giving speeches. 
There are also differences across the political spectrum, as can be seen in \autoref{fig:political-leaning}, with the right focusing much more on the \textit{security \& defense} framing compared to sources from the left or center across both modalities. 
To highlight this further, we show the finer-grained issue specific frames per political leaning in \autoref{fig:issue-frame-leaning}. 
When looking at frame frequencies normalised by each political leaning, we see clear differences and fine-grained signals about which specific framing publishers leaning to different sides of the political spectrum use. 
The left and center tend to highlight the \textit{humanitarian crisis} framing much more, the gap is smaller for right leaning publishers, while on the right, the focus is more on immigrants being an \textit{economic burden} or a \textit{national security threat}. 

\section{Discussion}

Being exposed to selective information can significantly bias our world view. For news, it can lead to problems at an individual level such as miscommunication; at a societal level, it can lead to misinformation, political polarisation, or avoidance of news~\citep{lecheler-framing-effects, iyengar2017typology}. This problem is aggravated considering news consumers tend to receive their news from social media, further creating ``curation bubbles''~\citep{GREEN_MCCABE_SHUGARS_CHWE_HORGAN_CAO_LAZER_2025}, where the dynamics of the networks on social media further contribute to curation, and in effect, exposure to selective partisan information. Thus, it is important to conduct analyses of news framing at scale to understand its effects~\citep{lecheler-framing-effects}.

\rev{Restricting computational framing analysis to text or single labels overlooks important information, as articles often contain multiple frames, and images might convey ideological or emotional undertones that text alone may miss \cite{wessler2016global,geise2025news}. While analysing framing in images is subjective, we find it possible to reach meaningful agreement among trained annotators and detect frames with reasonable accuracy at scale.}

Our analysis of multimodal framing in the news reveals that more frames are predicted for texts compared to images, that there are distinct lexical patterns associated with text frames vs. image frames, and that frames can be used for detailed topic analysis, including both at a corpus level and at a more fine-grained level by focusing on a single issue.~\rev{We show how news publishers across the political spectrum use different issue-specific frames to shape narratives, potentially reinforcing echo chambers and influencing misinformation and societal polarisation. The differences we uncover between the use of text and image frames emphasize that it is crucial to take images into account when analysing framing in the news.}

\rev{The cause of these different framings in images compared to the text is multi-faceted as they stem from multiple factors, including reliance on copyright-free images, algorithmic bias in image search, editorial intent, individual and institutional biases, and corporate ownership structures. Prior work classifies this as frame-building which shapes how framing appears in news ~\citep{de2005news}. Our method allows for an analysis of the outcome of this process, allowing for a more thorough investigation and analysis at scale.}

\section{Conclusion}

Framing analysis is an important task for content analysis at large, with implications for several fields in the social sciences, particularly for media bias analysis. In this work, we propose a methodology for leveraging open-weight large (vision-) language models to conduct integrative framing analysis of news articles at scale. While models are highly sensitive to prompting and over-rely on surface level visual features, we show that they can be be prompted to have reasonable accuracy on human-annotated framing datasets. Our approach allows for efficient, customisable, and replicable analyses of images and text in large-scale corpora. We demonstrate the usefulness of our approach by conducting multi-faceted framing analyses, taking into account both generic and issue-specific frames, textual and visual framing, and multiple framings per article. We find stark differences in visual and textual framing of the same topics, as well as differences in framing across publications across the political spectrum. We release a large scale dataset of 500k news articles with model annotations for article text as well as images, which we hope is useful for future studies in framing analysis.

\section{Limitations}

Despite the advancements in automating multimodal framing analysis, our approach has certain limitations. Firstly, our method assumes that both textual and visual elements contribute meaningfully to framing, yet in some cases, framing might be driven predominantly by one modality. This is akin to a news reader focusing more on one modality while consuming news. The challenge of evaluating the dominance of one modality remain an open problem when text and image signals conflict or reinforce different interpretations. \rev{Similarly, our approach also assumes equivalence across several frame predictions within an article i.e. all predicted frames have the same salience, however, in reality, one frame might be more dominant over another within the text or the image. We chose this design due to method constraints, accurately predicting multiple frames in an article or an image is already an very hard problem, detecting their relative salience is much harder. We ran some initial experiments with naively prompting for predictions in order of saliency but found the outputs to be quite noisy while doing qualitative analysis. Further, the outputs were hard to assess systematically since no such dataset with dominance scores exists for frame prediction.} Another limitation is that the interpretability of automated framing predictions remains limited. While our approach can identify patterns aligned with theoretical framing literature, it does not fully replicate the depth of qualitative human analysis. Ensuring transparency in model decisions and improving explainability remains an important avenue for future research. Finally, our study focuses on a specific set of framing methodologies and datasets, meaning generalizability to other media contexts or platforms requires further investigation. The dynamic nature of framing, influenced by evolving cultural and political landscapes, suggests that models must be continuously updated to remain effective.

\section{Ethical Considerations}
The models used in this study have been trained on large-scale datasets that may encode existing societal biases, which can impact the fairness and accuracy of framing predictions. If left uninvestigated, these biases may reinforce dominant narratives while marginalizing alternative perspectives, particularly in politically sensitive or socially divisive topics. Though having a scalable and automated approach can help aid news organizations and journalism scholars, it must be kept in mind that the generated responses are based on probabilistic patterns, hence we should allow error margins for some misclassifications. News is dynamic in nature but models can become static in nature, hence any method for news framing necessitates continuous updates of these large models.

\bibliography{custom}

\appendix

\section{Source selection}
\label{sec:source_domains}
We provide the list of all sources along with their corresponding political leanings in~\autoref{tab:source_domains}.
\begin{table}[]
    \footnotesize
    \centering
    \begin{tabular}{l>{\raggedright\arraybackslash}p{5cm}}
        \toprule
        Leaning & News Domain \\
        \midrule
        Left & alternet.org, editor.cnn.com, democracynow.org, dailybeast.com, huffpost.com, theintercept.com, jacobin.com, motherjones.com, newyorker.com, slate.com, msnbc.com, vox.com \\
        
        Left Leaning  &  abcnews.com, apnews.com, theatlantic.com, bloomberg.com, cbsnews.com, insider.com, nbcnews.com, thenytimes.com, npr.com, politico.com, propublica.org, time.com, washingtonpost.com, yahoonews.com, usatoday.com, theguardian.com \\
        
        Center & axios.com, forbes.com, newsweek.com, reuters.com, realclearpolitics.com, thehill.com \\
        
        Right Leaning & thedispatch.com, theepochtimes.com, foxbusiness.com, ijr.com, nypost.com, thepostmillennial.com, washingtonexaminer.com, washingtontimes.com \\
        
        Right & theamericanconservative.com, theamericanspectator.com, breitbart.com, dailycaller.com, dailywire.com, foxnews.com, newsmax.com, oann.com, thefederalist.com \\
        
        \bottomrule
    \end{tabular}
\caption{List of news sources used for our dataset split by their political leaning}
\label{tab:source_domains}
\end{table}

\section{Extracted Aspects}
\label{app:extracted_aspects}

\rev{In \autoref{tab:extracted_aspects}, we provide a list of extracted aspects per modality. For the main subject of the articles, the models were instructed to extract entities central to the text or the image, if there is one. We extract captions from the images and release it as part of the dataset but do not use it for our subsequent framing analysis.}

\begin{table}[t]
    \centering
    \footnotesize
    \begin{tabular}{p{1cm}|p{1.7cm}|p{3.5cm}}
    \toprule
       Modality &  Aspect  &  Description \\
       \midrule
       \multirow{ 5}{*}{Text} & 
       Topic  &  Broad topic of the article\\[0.5em]
       \cline{2-3}
       & Main subject & Subject of the text\\[0.5em]
    \cline{2-3}
       &Generic Frame & One or more of the 15 generic media frames\\[1em]
       \cline{2-3}
       & Issue Frame & Inductive, open-ended framing of a topic \\
       \midrule
        \multirow{ 5}{*}{Image} 
       &Caption &  Caption of the image\\[0.5em]
       \cline{2-3}
       &Main Subject & Subject in the image\\[0.5em]
       \cline{2-3}
       &Generic Frame & One or more of the 15 generic media frames\\
       \bottomrule
    \end{tabular}
    \caption{List of aspects extracted per modality.}
    \label{tab:extracted_aspects}
\end{table}

\begin{table}[]
\begin{tabular}{llll}
\toprule
Label & Precision & Recall & F1-score \\
\midrule
cap\&res      & 0.39   & 0.34     & 0.36 \\
crime         & 0.50   & 0.87     & 0.63 \\
culture       & 0.38   & 0.37     & 0.37 \\
economic      & 0.43   & 0.69     & 0.53 \\
fairness      & 0.17   & 0.74     & 0.28 \\
health        & 0.48   & 0.48     & 0.48 \\
legality      & 0.53   & 0.87     & 0.66 \\
morality      & 0.30   & 0.63     & 0.41 \\
policy        & 0.40   & 0.73     & 0.51 \\
political     & 0.68   & 0.53     & 0.60 \\
public\_op    & 0.32   & 0.55     & 0.40 \\
quality\_life & 0.28   & 0.36     & 0.31 \\
regulation    & 0.26   & 0.48     & 0.34 \\
security      & 0.30   & 0.45     & 0.36 \\
\midrule
micro avg     & 0.42   & 0.62     & 0.50 \\
macro avg     & 0.39   & 0.58     & 0.45 \\
weighted avg  & 0.45   & 0.62     & 0.51 \\
samples avg   & 0.46   & 0.63     & 0.51 \\
\bottomrule
\end{tabular}
\caption{Metrics per frame label (multi-label) for our text frame classifier on the MFC dataset}
\label{tab:text-frame-perf-label}
\end{table}

\begin{table*}[t]
    \centering
    \resizebox{1\linewidth}{!}{
    \scriptsize
    \begin{tabular}{@{}cclclclcl@{}}
    
        \toprule
         & \multicolumn{2}{c}{legality} & \multicolumn{2}{c}{morality} & \multicolumn{2}{c}{security} & \multicolumn{2}{c}{regulation} 
        \\ 
        \cmidrule(lr){2-3}\cmidrule(lr){4-5}\cmidrule(lr){6-7}\cmidrule(lr){8-9}
        & $z$-score & bigram & $z$-score & bigram & $z$-score & bigram & $z$-score & bigram 
        \\ 
        \midrule 
        \\ [-10pt]     
        
\multirow{10}{1.5cm}{\centering more associated with \textbf{images}}  & $\gradient{7.6}$ & supreme court & $\gradient{4.7}$ & pope francis & $\gradient{5.7}$ & air force & $\gradient{6.6}$ & secretary general \\
 & $\gradient{6.3}$ & bankman fried & $\gradient{4.1}$ & san francisco & $\gradient{3.2}$ & follow twitter & $\gradient{5.3}$ & united arab \\
 & $\gradient{5.6}$ & hunter biden & $\gradient{3.7}$ & supreme court & $\gradient{3.1}$ & border patrol & $\gradient{5.3}$ & arab emirates \\
 & $\gradient{5.1}$ & years prison & $\gradient{3.3}$ & new jersey & $\gradient{3.1}$ & fire department & $\gradient{4.9}$ & biden said \\
 & $\gradient{4.8}$ & attorney office & $\gradient{3.1}$ & two years & $\gradient{2.8}$ & united airlines & $\gradient{4.9}$ & prime minister \\
 & $\gradient{4.4}$ & clarence thomas & $\gradient{3.1}$ & law enforcement & $\gradient{2.6}$ & safety board & $\gradient{4.8}$ & join nato \\
 & $\gradient{4.2}$ & justice department & $\gradient{3.0}$ & family members & $\gradient{2.5}$ & climate change & $\gradient{4.3}$ & saudi arabia \\
 & $\gradient{4.1}$ & district attorney & $\gradient{2.9}$ & catholic church & $\gradient{2.5}$ & federal government & $\gradient{4.2}$ & official said \\
 & $\gradient{4.1}$ & rights act & $\gradient{2.8}$ & high school & $\gradient{2.4}$ & russian forces & $\gradient{4.1}$ & general jens \\
 & $\gradient{3.9}$ & high school & $\gradient{1.9}$ & took place & $\gradient{2.4}$ & natural gas & $\gradient{4.1}$ & jens stoltenberg \\
 
\midrule 

\multirow{10}{1.5cm}{\centering more associated with \textbf{texts}}  & $\gradient{-2.8}$ & north korea & $\gradient{-2.1}$ & president donald & $\gradient{-2.8}$ & debt limit & $\gradient{-0.8}$ & donald trump \\
 & $\gradient{-2.8}$ & trump said & $\gradient{-2.1}$ & president biden & $\gradient{-2.9}$ & attempted murder & $\gradient{-0.8}$ & officials said \\
 & $\gradient{-2.9}$ & loan forgiveness & $\gradient{-2.1}$ & vice president & $\gradient{-3.3}$ & anyone information & $\gradient{-0.9}$ & president biden \\
 & $\gradient{-2.9}$ & interest rates & $\gradient{-2.1}$ & social media & $\gradient{-3.3}$ & york city & $\gradient{-1.1}$ & last week \\
 & $\gradient{-3.0}$ & gov ron & $\gradient{-2.1}$ & joe biden & $\gradient{-3.4}$ & old man & $\gradient{-1.1}$ & news app \\
 & $\gradient{-3.2}$ & nbc news & $\gradient{-2.1}$ & request comment & $\gradient{-3.5}$ & san francisco & $\gradient{-1.1}$ & click get \\
 & $\gradient{-3.3}$ & ron desantis & $\gradient{-2.5}$ & white house & $\gradient{-3.7}$ & hong kong & $\gradient{-1.2}$ & get fox \\
 & $\gradient{-3.4}$ & prime minister & $\gradient{-3.1}$ & donald trump & $\gradient{-5.1}$ & police department & $\gradient{-1.2}$ & united states \\
 & $\gradient{-3.6}$ & new hampshire & $\gradient{-3.2}$ & fox news & $\gradient{-5.3}$ & year old & $\gradient{-1.6}$ & new york \\
 & $\gradient{-3.6}$ & national security & $\gradient{-3.2}$ & former president & $\gradient{-7.5}$ & police said & $\gradient{-2.1}$ & fox news \\
 
        \bottomrule \\

        & \multicolumn{2}{c}{culture} & \multicolumn{2}{c}{fairness} & \multicolumn{2}{c}{health} & \multicolumn{2}{c}{public opinion} 
        \\ 
        \cmidrule(lr){2-3}\cmidrule(lr){4-5}\cmidrule(lr){6-7}\cmidrule(lr){8-9}
        & $z$-score & bigram & $z$-score & bigram & $z$-score & bigram & $z$-score & bigram 
        \\ 
        \midrule 
        \\ [-10pt]     
        
\multirow{10}{1.5cm}{\centering more associated with \textbf{images}}  & $\gradient{8.1}$ & taylor swift & $\gradient{5.8}$ & los angeles & $\gradient{5.2}$ & long term & $\gradient{6.5}$ & supreme court \\
 & $\gradient{7.3}$ & eras tour & $\gradient{5.2}$ & student debt & $\gradient{5.1}$ & term care & $\gradient{5.9}$ & sag aftra \\
 & $\gradient{6.1}$ & getty images & $\gradient{5.1}$ & monthly payments & $\gradient{4.9}$ & care insurance & $\gradient{5.0}$ & artificial intelligence \\
 & $\gradient{5.2}$ & las vegas & $\gradient{4.8}$ & san francisco & $\gradient{4.3}$ & year year & $\gradient{4.5}$ & biden administration \\
 & $\gradient{4.9}$ & swift eras & $\gradient{4.8}$ & fain said & $\gradient{4.2}$ & weight loss & $\gradient{4.3}$ & anti israel \\
 & $\gradient{4.6}$ & box office & $\gradient{4.4}$ & student loan & $\gradient{3.9}$ & long covid & $\gradient{4.0}$ & cbs news \\
 & $\gradient{4.2}$ & los angeles & $\gradient{4.3}$ & auto workers & $\gradient{3.5}$ & five year & $\gradient{4.0}$ & san francisco \\
 & $\gradient{4.1}$ & fourth july & $\gradient{4.3}$ & general motors & $\gradient{3.4}$ & prime day & $\gradient{3.8}$ & auto workers \\
 & $\gradient{3.7}$ & instagram post & $\gradient{4.1}$ & writers strike & $\gradient{3.0}$ & health insurance & $\gradient{3.8}$ & mar lago \\
 & $\gradient{3.6}$ & last year & $\gradient{4.1}$ & official said & $\gradient{2.8}$ & disease control & $\gradient{3.8}$ & big three \\
 
\midrule 

\multirow{10}{1.5cm}{\centering more associated with \textbf{texts}}  & $\gradient{-3.0}$ & ads content & $\gradient{-2.7}$ & south carolina & $\gradient{-2.4}$ & former president & $\gradient{-2.9}$ & state police \\
 & $\gradient{-3.0}$ & parties information & $\gradient{-2.7}$ & nikki haley & $\gradient{-2.5}$ & county sheriff & $\gradient{-3.2}$ & police department \\
 & $\gradient{-3.0}$ & charities online & $\gradient{-2.9}$ & special counsel & $\gradient{-2.5}$ & tropical storm & $\gradient{-3.2}$ & eras tour \\
 & $\gradient{-3.0}$ & contain info & $\gradient{-3.3}$ & ron desantis & $\gradient{-2.5}$ & medical examiner & $\gradient{-3.3}$ & social media \\
 & $\gradient{-3.0}$ & information see & $\gradient{-3.4}$ & attorney general & $\gradient{-2.6}$ & cause death & $\gradient{-3.5}$ & window share \\
 & $\gradient{-3.0}$ & online ads & $\gradient{-3.6}$ & hunter biden & $\gradient{-2.7}$ & cbs essentials & $\gradient{-3.7}$ & new window \\
 & $\gradient{-3.0}$ & newsletters may & $\gradient{-3.7}$ & president donald & $\gradient{-2.8}$ & sheriff office & $\gradient{-3.7}$ & opens new \\
 & $\gradient{-3.8}$ & young people & $\gradient{-3.9}$ & new hampshire & $\gradient{-3.1}$ & mother day & $\gradient{-4.3}$ & new hampshire \\
 & $\gradient{-3.8}$ & new hampshire & $\gradient{-5.4}$ & donald trump & $\gradient{-3.4}$ & office said & $\gradient{-4.5}$ & bud light \\
 & $\gradient{-4.2}$ & supreme court & $\gradient{-5.4}$ & former president & $\gradient{-5.0}$ & year old & $\gradient{-5.3}$ & taylor swift \\
 
        \bottomrule \\

        & \multicolumn{2}{c}{policy} & \multicolumn{2}{c}{capacity \& resources} & \multicolumn{2}{c}{political} & \multicolumn{2}{c}{economic} 
        \\ 
        \cmidrule(lr){2-3}\cmidrule(lr){4-5}\cmidrule(lr){6-7}\cmidrule(lr){8-9}
        & $z$-score & bigram & $z$-score & bigram & $z$-score & bigram & $z$-score & bigram 
        \\ 
        \midrule 
        \\ [-10pt]     
        
\multirow{10}{1.5cm}{\centering more associated with \textbf{images}}  & $\gradient{6.9}$ & white house & $\gradient{5.1}$ & climate change & $\gradient{10.2}$ & former president & $\gradient{13.1}$ & credit card \\
 & $\gradient{6.6}$ & hunter biden & $\gradient{5.0}$ & year old & $\gradient{7.4}$ & donald trump & $\gradient{12.2}$ & personal loan \\
 & $\gradient{6.3}$ & ron desantis & $\gradient{4.9}$ & officials said & $\gradient{6.8}$ & student loan & $\gradient{12.2}$ & credit score \\
 & $\gradient{6.0}$ & house republicans & $\gradient{4.2}$ & news app & $\gradient{6.5}$ & white house & $\gradient{12.2}$ & interest rates \\
 & $\gradient{5.8}$ & mortgage rates & $\gradient{4.1}$ & click get & $\gradient{6.3}$ & president donald & $\gradient{11.5}$ & interest rate \\
 & $\gradient{5.6}$ & president biden & $\gradient{4.1}$ & oil gas & $\gradient{5.8}$ & continue reading & $\gradient{10.6}$ & card debt \\
 & $\gradient{5.3}$ & house speaker & $\gradient{3.9}$ & fox news & $\gradient{5.4}$ & debt ceiling & $\gradient{9.0}$ & social security \\
 & $\gradient{5.1}$ & kevin mccarthy & $\gradient{3.7}$ & said statement & $\gradient{5.3}$ & supreme court & $\gradient{8.1}$ & lower interest \\
 & $\gradient{5.0}$ & prime minister & $\gradient{3.6}$ & united states & $\gradient{5.0}$ & joe biden & $\gradient{8.0}$ & high yield \\
 & $\gradient{5.0}$ & donald trump & $\gradient{3.5}$ & supreme court & $\gradient{5.0}$ & primary ballot & $\gradient{7.7}$ & yield savings \\
 
\midrule 

\multirow{10}{1.5cm}{\centering more associated with \textbf{texts}}  & $\gradient{-3.1}$ & recent years & $\gradient{-3.3}$ & card debt & $\gradient{-3.3}$ & taylor swift & $\gradient{-2.3}$ & news digital \\
 & $\gradient{-3.1}$ & social security & $\gradient{-3.3}$ & wall street & $\gradient{-3.5}$ & police officer & $\gradient{-2.4}$ & click get \\
 & $\gradient{-3.2}$ & told cbs & $\gradient{-3.4}$ & savings accounts & $\gradient{-3.6}$ & city council & $\gradient{-2.4}$ & climate change \\
 & $\gradient{-3.2}$ & per month & $\gradient{-3.6}$ & cash flow & $\gradient{-3.8}$ & san francisco & $\gradient{-2.5}$ & news app \\
 & $\gradient{-3.2}$ & police said & $\gradient{-3.8}$ & federal reserve & $\gradient{-4.2}$ & los angeles & $\gradient{-2.5}$ & former president \\
 & $\gradient{-3.3}$ & los angeles & $\gradient{-4.5}$ & long term & $\gradient{-4.3}$ & police department & $\gradient{-2.6}$ & social media \\
 & $\gradient{-3.4}$ & officials said & $\gradient{-4.5}$ & credit score & $\gradient{-4.3}$ & window share & $\gradient{-2.6}$ & white house \\
 & $\gradient{-3.5}$ & border patrol & $\gradient{-4.7}$ & interest rate & $\gradient{-4.4}$ & social media & $\gradient{-2.6}$ & loan forgiveness \\
 & $\gradient{-4.3}$ & mental health & $\gradient{-5.0}$ & credit card & $\gradient{-4.6}$ & new window & $\gradient{-2.8}$ & privacy policy \\
 & $\gradient{-4.9}$ & year old & $\gradient{-6.0}$ & interest rates & $\gradient{-4.6}$ & opens new & $\gradient{-3.7}$ & fox news \\
 
        \bottomrule \\
    \end{tabular}
    
    }
    \caption{
        The bigrams most associated the \textbf{images} (higher $z$-scores) and \textbf{texts} (lower $z$-scores) for all the frames except crime and quality of life (shown in Table \ref{table:fightin-words-condensed}).
    }
    \label{table:fightin-words-full-2}
\end{table*}

\section{Data Filtering}
\label{app:filtering}
The scraped data carried a lot of noise. There were single line or extremely long articles, images that only depicted logos of news websites, which are not useful for analysis. To remove these, we filtered the articles whose lengths below the 5th percentile and above the 95th percentile, removing the outliers. \rev{We also removed articles that were not in English, an information available via news-please library.
For images, we similarly removed the files for which the size was above the 95th percentile. There were several images in the dataset that were only logos of news organisations due to the scraper picking the wrong image from the webpage. These only had the logo and no other content, making them irrelevant for our framing analysis. There was no simple way to remove these images, we approximated their identification based on the image size. We distributed the images per their size into bins and plotted 50 randomly sampled images from them. After qualitatively assessing images from each bin, we concluded that 90\% of the visualised images below the 5000 bytes threshold mostly constituted of logos only, while it was much lower for higher threshold. Therefore, we remove all figures with a size of less than 5000 bytes.}

\section{Frame Descriptions}
\label{app:frame_desc}
In \autoref{tab:frame_desc}, we provide the names and corresponding descriptions for the generic frames used in our work. There are minor differences in the frame names provided here and the ones used in our dataset. This was due to us relying on the version released by~\citet{card2015media} with the Media Frames Corpus dataset, which builds on the same dimensions. 

\section{Experimental Details}
\label{app:model-exp-details}

\textbf{Mistral}: We used Mistral-7B-Instruct-v0.3 available via \rev{huggingface} \footnote{https://huggingface.co/mistralai/Mistral-7B-v0.3}. We used vLLM for high-throughput and memory-efficient inference and set the parameters as temperature=0.2, max\_tokens=4000, dtype='half' and  max\_model\_len=8096. \\

\noindent \textbf{Pixtral}: We used Pixtral-12B-2409 available via hugging face \footnote{https://huggingface.co/mistralai/Pixtral-12B-2409}. Similar to Mistral, we used vLLM and set the parameters as: temperature=0.2, max\_tokens=1024, dtype=`half' and max\_model\_len=7000. Before processing the images via pixtral, we also resized the images to 512 x 512 for computational efficiency. 

For vision annotations, we ran pixtral on 8 Nvidia A100s which enabled us to finish the computation in 5 days. For text annotations, we ran mistral on a mix of Nvidia TitanRTX and A100s and finished the computation in similar time as vision annotations. 

\begin{table*}[h]
    \centering
    \footnotesize
    \begin{tabular}{lp{10cm}}
        \toprule
        \textbf{Frame Name} & \textbf{Description} \\
        \midrule
        Economic  & The costs, benefits, or monetary/financial implications of the issue (to an individual, family, community or to the economy as a whole) \\
        \midrule
        Capacity and resources  &  The lack of or availability of physical, geographical, spatial, human, and financial resources, or the capacity of existing systems and resources to implement or carry out policy goals.\\
        \midrule
        Morality  & Any perspective—or policy objective or action (including proposed action)—that is compelled by religious doctrine or interpretation, duty, honor, righteousness or any other sense of ethics or social responsibility.\\
        \midrule
        Fairness and equality  &  Equality or inequality with which laws, punishment, rewards, and resources are applied or distributed among individuals or groups. Also the balance between the rights or interests of one individual or group compared to another individual or group.\\
        \midrule
        Constitutionality and jurisprudence  & The constraints imposed on or freedoms granted to individuals, government, and corporations via the Constitution, Bill of Rights and other amendments, or judicial interpretation. This deals specifically with the authority of government to regulate, and the authority of individuals/corporations to act independently of government.\\
        \midrule
        Policy prescription and evaluation & Particular policies proposed for addressing an identified problem, and figuring out if certain policies will work, or if existing policies are effective.\\
        \midrule
        Law and order, crime and justice  & Specific policies in practice and their enforcement, incentives, and implications. Includes stories about enforcement and interpretation of laws by individuals and law enforcement, breaking laws, loopholes, fines, sentencing and punishment. Increases or reductions in crime.\\
        \midrule
        Security and defence  & Security, threats to security, and protection of one’s person, family, in-group, nation, etc. Generally an action or a call to action that can be taken to protect the welfare of a person, group, nation sometimes from a not yet manifested threat.\\
        \midrule
        Health and safety  & Healthcare access and effectiveness, illness, disease, sanitation, obesity, mental health effects, prevention of or perpetuation of gun violence, infrastructure and building safety.\\
        \midrule
        Quality of life  & The effects of a policy on individuals’ wealth, mobility, access to resources, happiness, social structures, ease of day-to-day routines, quality of community life etc. \\
        \midrule
        Cultural identity  & The social norms, trends, values and customs constituting culture(s), as they relate to a specific policy issue\\
        \midrule
        Public opinion  & References to general social attitudes, polling and demographic information, as well as implied or actual consequences of diverging from or getting ahead of public opinion or polls.\\
        \midrule
        Political  & Any political considerations surrounding an issue. Issue actions or efforts or stances that are political, such as partisan filibusters, lobbyist involvement, bipartisan efforts, deal-making and vote trading, appealing to one’s base, mentions of political manoeuvring. Explicit statements that a policy issue is good or bad for a particular political party.\\
        \midrule
        External regulation and reputation  & The United States’ external relations with another nation; the external relations of one state with another; or relations between groups. This includes trade agreements and outcomes, comparisons of policy outcomes or desired policy outcomes.\\
        \midrule
        Other  & Any frames that do not fit into the above categories.\\
        \bottomrule
    \end{tabular}
    \caption{Frame name and descriptions for each frame used in the dataset}
    \label{tab:frame_desc}
\end{table*}

\section{Article Subject Portrayal}

\begin{figure}
    \centering
    \includegraphics[width=0.8\linewidth]{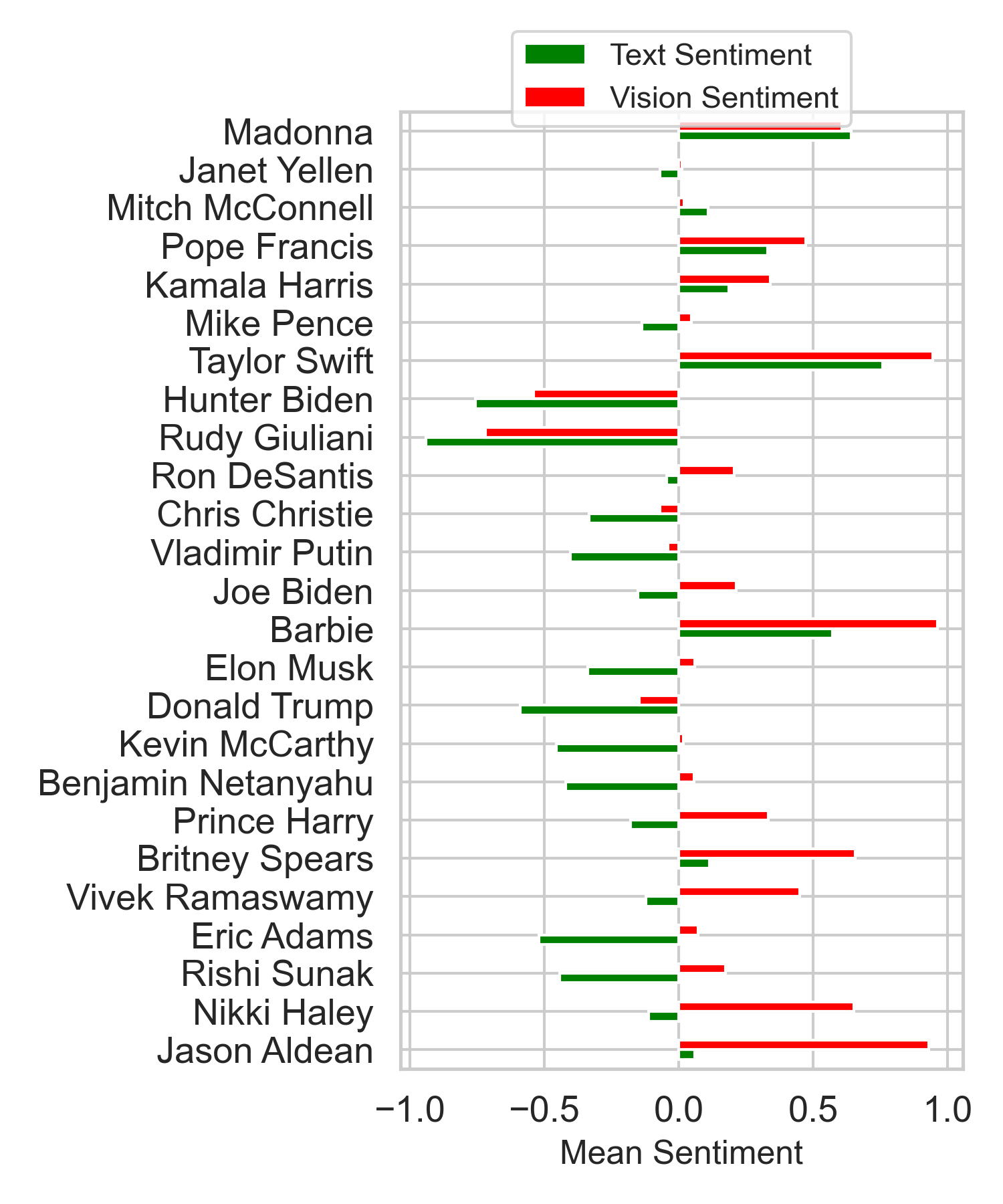}
    \caption{Difference in sentiment of the main subject's portrayal across text and images}
    \label{fig:entt_sentiment}
\end{figure}

We additionally explore how individual entities in the articles are portrayed. For this analysis, we take the subset where the article and the image are portraying the same main entity. The entities and sentiment are extracted by prompting the LLM and VLM to identify the main subject in the article or the image, as shown in the prompt in \autoref{lst:prompt-image-frame-prediction}.
\autoref{fig:entt_sentiment} shows peculiar differences when contrasting the image portrayal from the portrayal in the articles. 
A general pattern that can be observed is that entities were portrayed more positively in the images, compared to the text, e.g., Rishi Sunak, Prince Harry, Elon Musk, Joe Biden, Benjamin Netanyahu. 
For some, there is a negative portrayal overall that is much more explicit in the text, e.g., Donald Trump, Hunter Biden, Vladimir Putin, Rudy \rev{Giuliani}. 

\section{Frame Classifier Error Analysis}
\label{app:frame-class-perf}
\rev{\paragraph{Text} We provide the per label performance of the framing classifier for text on the MFC dataset in \autoref{tab:text-frame-perf-label}}

\rev{We additionally show examples of misclassifications on the dataset in \autoref{tab:mfc_misclassification}. Here, while the intersection of the prediction and label sets is null, the explanation of the model for identifying those framings is reasonable and faithful to the article text.
To understand patterns in these misclassifications, we analyse labels that the model often gets confused with. Since we’re operating in a multi-label setting, a traditional confusion matrix cannot be applied. A multi-label confusion matrix also treats the labels as binary and not give insight into confusion patterns. To provide insight into these specific confusions, we calculate a mismatch frequency matrix where for each missed ground truth label in an instance (rows in the table below), we mark the erroneous predictions as 1 (columns). We then sum these for the entire dataset, giving us an overview of which labels are often mis-predicted for each missed gold label. For the text frames, the mismatch frequency against MFC is shown in \autoref{tab:text_mismatch_matrix}. The most frequent mismatch is with the \textit{political} frame, with the model predicting \textit{legality, policy} or \textit{crime} framing instead. \textit{Culture} is another frame often mislabeled, albeit with much lower frequencies.}

\rev{ \paragraph{Image} Similarly, for the image error analysis, we calculate the mismatch frequency matrix as shown in \autoref{tab:vision_mismatch_matrix}. As can be seen, most mispredictions involve the \textit{None} label, with the model over or underpredicting it. Other labels often confused are \textit{public opinion} and \textit{policy prescription}, the latter being particularly hard to detect in images even during human annotations.}

\rev{Examining the ~15\% of examples where the model's predicted frames had no overlap with the human-labeled (gold-standard) frames, and we will add this error analysis to the paper. Among these zero-overlap cases, the most common human-assigned label was `none', which accounted for 40.9\% of the instances. This indicates that in many such cases, the model over-predicted by assigning frames even when annotators found no meaningful framing. The most frequently missed substantive frame was `public opinion' (11.8\%), followed by `cultural identity' (10.8\%). Other frames commonly missed included `quality of life', `political', `capacity and resources', and `health and safety', each appearing in approximately 7.5\% of the zero-overlap cases. Less frequently missed frames were `external regulation and reputation' (3.2\%), and `economic', `morality', and `crime and punishment', each at 1.1%
We also note that some frames were relatively infrequent in the human-labeled data overall. For instance, `fairness and equality' appeared in only 4.1\% of annotated examples, followed by `morality' (2.31\%), `external regulation and reputation' (2.31\%), `economic' (1.92\%), `legality, constitutionality and jurisprudence' (1.92\%), and `policy prescription and evaluation' (1.54\%). As a result, their low miss rates may reflect either their lower frequency in the dataset or higher accuracy in the model’s predictions.}

\begin{table}[]
\begin{adjustbox}{angle=90}
\centering
\small
\begin{tabular}{lrrrrrrrrrrrrrrr}
\textbf{} &
  \textbf{none} &
  \textbf{economic} &
  \textbf{cap\&res} &
  \textbf{morality} &
  \textbf{fairness} &
  \textbf{legality} &
  \textbf{policy} &
  \textbf{crime} &
  \textbf{security} &
  \textbf{health} &
  \textbf{quality\_life} &
  \textbf{culture} &
  \textbf{public\_op} &
  \textbf{political} &
  \textbf{regulation} \\
none          & 0  & 0    & 0   & 0   & 0    & 0    & 0    & 0    & 0    & 0   & 0    & 0   & 0    & 0   & 0   \\
economic      & 5  & 0    & 6   & 198 & 724  & 897  & 970  & 798  & 167  & 178 & 152  & 152 & 408  & 261 & 180 \\
cap\&res      & 20 & 1205 & 0   & 95  & 438  & 953  & 1497 & 380  & 349  & 913 & 854  & 213 & 443  & 140 & 309 \\
morality      & 2  & 207  & 55  & 0   & 340  & 318  & 392  & 282  & 73   & 103 & 164  & 163 & 235  & 150 & 69  \\
fairness      & 0  & 80   & 20  & 120 & 0    & 177  & 246  & 189  & 37   & 32  & 55   & 88  & 139  & 148 & 29  \\
legality      & 2  & 296  & 153 & 421 & 664  & 0    & 479  & 329  & 182  & 100 & 165  & 285 & 420  & 265 & 100 \\
policy        & 12 & 282  & 126 & 567 & 1262 & 1042 & 0    & 649  & 210  & 175 & 308  & 426 & 424  & 478 & 132 \\
crime         & 1  & 204  & 72  & 158 & 498  & 364  & 387  & 0    & 98   & 64  & 80   & 164 & 268  & 349 & 77  \\
security      & 1  & 341  & 108 & 117 & 524  & 866  & 809  & 806  & 0    & 60  & 183  & 148 & 397  & 276 & 182 \\
health        & 8  & 541  & 194 & 435 & 871  & 1065 & 1218 & 1074 & 507  & 0   & 448  & 300 & 669  & 341 & 194 \\
quality\_life & 8  & 639  & 175 & 563 & 1462 & 713  & 1181 & 445  & 197  & 237 & 0    & 563 & 892  & 505 & 181 \\
culture       & 15 & 797  & 241 & 576 & 1036 & 1189 & 1766 & 1179 & 390  & 438 & 634  & 0   & 897  & 563 & 319 \\
public\_op    & 4  & 278  & 77  & 383 & 954  & 951  & 796  & 846  & 219  & 127 & 289  & 364 & 0    & 406 & 142 \\
political     & 21 & 1529 & 406 & 535 & 1944 & 3172 & 3199 & 2308 & 1143 & 824 & 1136 & 569 & 1243 & 0   & 421 \\
regulation    & 6  & 401  & 62  & 52  & 220  & 399  & 529  & 205  & 184  & 168 & 221  & 108 & 179  & 79  & 0  
\end{tabular}
\end{adjustbox}
\caption{\rev{Mismatch frequency matrix for the text annotations on the MFC dataset}}
\label{tab:text_mismatch_matrix}
\end{table}

\begin{table}[]
\centering
\small
\begin{adjustbox}{angle=90}
\begin{tabular}{lrrrrrrrrrrrrrrr}
\textbf{}     & \textbf{cap\&res} & \textbf{crime} & \textbf{culture} & \textbf{economic} & \textbf{fairness} & \textbf{health} & \textbf{legality} & \textbf{morality} & \textbf{none} & \textbf{policy} & \textbf{political} & \textbf{public\_op} & \textbf{quality\_life} & \textbf{regulation} & \textbf{security} \\
cap\&res      & 0                 & 3              & 0                & 0                 & 0                 & 3               & 1                 & 1                 & 14            & 2               & 0                  & 4                   & 1                      & 0                   & 6                 \\
crime         & 0                 & 0              & 0                & 0                 & 0                 & 0               & 5                 & 0                 & 3             & 0               & 0                  & 1                   & 0                      & 0                   & 0                 \\
culture       & 2                 & 0              & 0                & 0                 & 4                 & 3               & 2                 & 3                 & 16            & 2               & 0                  & 3                   & 2                      & 0                   & 4                 \\
economic      & 0                 & 1              & 0                & 0                 & 0                 & 0               & 0                 & 0                 & 2             & 0               & 0                  & 0                   & 1                      & 0                   & 0                 \\
fairness      & 0                 & 0              & 0                & 0                 & 0                 & 0               & 2                 & 0                 & 1             & 1               & 0                  & 0                   & 0                      & 0                   & 0                 \\
health        & 2                 & 4              & 1                & 0                 & 0                 & 0               & 3                 & 3                 & 11            & 2               & 0                  & 4                   & 1                      & 0                   & 7                 \\
legality      & 0                 & 0              & 0                & 0                 & 0                 & 0               & 0                 & 0                 & 0             & 0               & 0                  & 0                   & 0                      & 0                   & 0                 \\
morality      & 0                 & 1              & 0                & 0                 & 0                 & 0               & 0                 & 0                 & 0             & 0               & 0                  & 0                   & 0                      & 0                   & 1                 \\
none          & 4                 & 3              & 7                & 0                 & 3                 & 4               & 9                 & 2                 & 0             & 21              & 4                  & 6                   & 4                      & 0                   & 8                 \\
policy        & 0                 & 0              & 0                & 0                 & 0                 & 0               & 1                 & 0                 & 0             & 0               & 0                  & 0                   & 0                      & 0                   & 0                 \\
political     & 0                 & 0              & 0                & 0                 & 1                 & 0               & 3                 & 2                 & 9             & 9               & 0                  & 0                   & 0                      & 1                   & 3                 \\
public\_op    & 1                 & 0              & 1                & 0                 & 0                 & 0               & 6                 & 0                 & 7             & 11              & 1                  & 0                   & 0                      & 1                   & 0                 \\
quality\_life & 3                 & 0              & 1                & 0                 & 2                 & 3               & 2                 & 3                 & 21            & 1               & 0                  & 2                   & 0                      & 0                   & 0                 \\
regulation    & 1                 & 0              & 0                & 0                 & 0                 & 0               & 1                 & 1                 & 2             & 2               & 0                  & 0                   & 0                      & 0                   & 3                 \\
security      & 0                 & 1              & 0                & 0                 & 0                 & 0               & 2                 & 0                 & 2             & 0               & 0                  & 1                   & 0                      & 0                   & 0                
\end{tabular}
\end{adjustbox}
    \caption{\rev{Mismatch Matrix for Vision annotations across the human annotated dataset}}
    \label{tab:vision_mismatch_matrix}
\end{table}

\begin{table*}[]
\small
\centering
\begin{tabular}{p{2cm}p{2cm}p{2cm}p{9cm}}
\toprule
Topic & Prediction  & Label & Model reasoning \\
\midrule
gun\_control & ['crime', 'policy', 'security'] & ['political', 'health'] & The article discusses a gun ban for people on the no-fly list, which is related to crime and punishment and policy prescription and evaluation. Additionally, the context of the discussion is in response to a security threat, making it relevant to the Security and defense frame. \\
climate\_change & ['economic', 'health', 'quality\_life', 'policy', 'regulation'] & ['cap\&res', 'political'] & The article discusses the warming of the oceans over the last 40 years, which has implications for economic costs (e.g., potential damage to marine life and industries), health and safety (e.g., potential impact on human health and marine ecosystems), quality of life (e.g., potential effects on the overall well-being of individuals and communities), policy prescription and evaluation (e.g., the study provides evidence for or against computer models used to predict Earth's warming), and external regulation and reputation (e.g., international implications of climate change and the role of the U.S. in addressing it). \\
gun\_control & ['economic', 'crime', 'policy', 'public\_op', 'culture'] & ['regulation', 'political', 'quality\_life'] & The article discusses the financial implications of gun violence (Economic), the legal and enforcement aspects of gun control (Crime and punishment), the proposed policy changes to address the issue (Policy prescription and evaluation), the public opinion and sentiments towards gun control (Public sentiment), and the cultural differences between Japan and America in terms of gun control (Cultural identity). \\
climate\_change & ['economic', 'security', 'quality\_life'] & ['cap\&res', 'health', 'culture'] & The article discusses the potential impact of global warming on Antarctica's ice, which could lead to a rise in sea levels around the world, affecting coastal cities and potentially causing economic, security, and quality of life issues. Additionally, the article mentions the climate change aspect related to the melting of Antarctica's ice. \\
Tobacco & ['health', 'legality', 'fairness'] & ['policy', 'economic', 'quality\_life'] & The article discusses the health implications of smoking, the legality of banning smoking in bars and restaurants, and the potential for unequal treatment of smokers compared to non-smokers. \\
climate\_change & ['economic', 'legality', 'policy'] & ['political', 'regulation', 'cap\&res'] & The article discusses the financial implications of adhering to the Kyoto treaty on global warming for the U.S. economy (Economic). The President's statement that he couldn't in good faith sign Kyoto because it does not apply to other nations implies a discussion of the treaty's legality and constitutionality (Legality, constitutionality and jurisprudence). The article also mentions the White House disputing the science behind the conclusion that greenhouse gas emissions contribute to global warming, which suggests a policy prescription and evaluation (Policy prescription and evaluation). \\
Immigration & ['economic', 'legality', 'fairness', 'policy', 'public\_op'] & ['quality\_life', 'morality', 'political'] & The article discusses the financial implications of cutting federal assistance to legal immigrants (Economic), the constitutionality of denying assistance to legal residents (Legality, constitutionality and jurisprudence), the balance or distribution of rights and resources for legal immigrants compared to other residents (Fairness and equality), the discussion of specific policies aimed at addressing the issue (Policy prescription and evaluation), and the public sentiment towards allowing elderly immigrants to starve or children to live on the street (Public sentiment). \\
climate\_change & ['economic', 'policy', 'legality', 'public\_op', 'political', 'regulation'] & ['culture', 'cap\&res', 'health'] & The article discusses the decision of a TV station manager to not cover global warming stories, citing financial implications (Economic), the evolution of the issue into politics (Policy prescription and evaluation, Political), the perceived lack of conclusive scientific evidence (Legality, constitutionality and jurisprudence), public sentiment towards the issue (Public sentiment), and international implications of the U.S.'s stance on global warming (External regulation and reputation). \\
Immigration & ['morality', 'political', 'culture', 'crime'] & ['public\_op', 'fairness', 'security'] & The article discusses the protest against a border patrol group, which is framed as racist and un-American, thus invoking moral and political implications. The protest is organized by a Latino civil rights organization, indicating cultural identity. The group's activities are framed as a form of policing (crime and punishment), as they plan to patrol the border and ensure immigrants follow certain procedures. \\
\bottomrule
\end{tabular}
\caption{Example misclassifications on the MFC dataset along with the model reasoning for the predictions}
    \label{tab:mfc_misclassification}
\end{table*}

\section{Issue Frame Analysis}
\label{app:issue_frame_analysis}

\rev{As outlined in \autoref{subsec:model_anno}, we extract open-ended issue specific frames by provided the model with example issue frames for some issues and instructed to identify and generate appropriate framing of the issue in the article, based on its topic. The generated outputs are limited to few words, but are open-ended as the model is not instructed to choose from a given set of frames but rather generates them. Issue frames are tailored towards fine-grained analysis of specific issues, so they do not repeat a lot. In total, our dataset has over 56k unique issue frames. Some, however, do repeat across topics, we show the top 20 predicted issue frames by frequency, along with the top 3 broad level topics across which they appear in \autoref{tab:top_issue_frames}. As is evident from the table, the overall frequencies are low in the context of the dataset. Most issue-frames are spread across multiple topics rather than being concentrated into one. Some exceptions include Political Crisis, Corruption, Manipulation, and Power Struggle, for which Politics is the major topic across which they appear, which is intuitive. Others like Public Safety Concern are also concentrated under Crime.}

\begin{table*}[]
    \centering
    \small
    \begin{tabular}{lrl}
\toprule
Issue Frame & Count & Top Topics \\
\midrule
Humanitarian Crisis & 1723 & [('War', 437), ('Politics', 284), ('Immigration', 272)] \\
Political Crisis & 1470 & [('Politics', 1071), ('Legal', 294), ('Immigration', 49)] \\
Public Health Crisis & 1176 & [('Health', 797), ('Environment', 155), ('Crime', 84)] \\
Political Persecution & 1134 & [('Politics', 606), ('Legal', 519), ('Crime', 2)] \\
Political Corruption & 1064 & [('Politics', 843), ('Legal', 207), ('Crime', 12)] \\
Public Safety Concern & 1053 & [('Crime', 732), ('Legal', 42), ('Safety', 37)] \\
Political Manipulation & 931 & [('Politics', 667), ('Legal', 179), ('Immigration', 44)] \\
Natural Disaster & 907 & [('Environment', 328), ('Weather', 253), ('Disaster', 105)] \\
National Security Threat & 824 & [('Politics', 375), ('War', 180), ('Immigration', 63)] \\
Political Power Struggle & 736 & [('Politics', 649), ('Legal', 60), ('War', 10)] \\
Cultural Celebration & 627 & [('Culture', 285), ('Entertainment', 178), ('no\_topic', 58)] \\
Political Scandal & 622 & [('Politics', 446), ('Legal', 166), ('Crime', 3)] \\
Security Threat & 604 & [('War', 244), ('Politics', 139), ('Immigration', 80)] \\
Natural Disaster Threat & 589 & [('Environment', 297), ('Weather', 112), ('Natural Disasters', 79)] \\
Economic Struggle & 587 & [('Economy', 364), ('Business', 124), ('Politics', 38)] \\
Economic Burden & 587 & [('Economy', 152), ('Business', 128), ('Immigration', 69)] \\
Financial Opportunity & 578 & [('Finance', 268), ('Business', 229), ('Economy', 48)] \\
Tragedy & 566 & [('Crime', 217), ('Accident', 108), ('Transportation', 46)] \\
Legal Battle & 558 & [('Legal', 460), ('Politics', 69), ('Crime', 11)] \\
Criminal Threat & 557 & [('Crime', 512), ('Immigration', 26), ('Legal', 9)] \\
\bottomrule
\end{tabular}
    \caption{Most frequent issue-frames across the dataset along with the top 3 article topics they are encountered in}
    \label{tab:top_issue_frames}
\end{table*}

\section{Additional Plots}

\begin{figure}
    \centering
    \begin{subfigure}{0.5\textwidth}
        \centering
        \includegraphics[width=\textwidth]{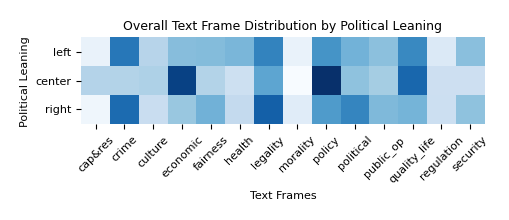}
    \end{subfigure}
    
    \vspace{0.1cm} %

    \begin{subfigure}{0.5\textwidth}
        \centering
        \includegraphics[width=\textwidth]{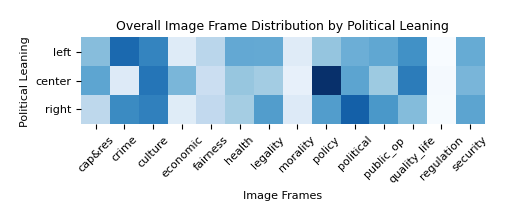}
    \end{subfigure}

    \caption{Overall comparison of text and image frame distributions across political leanings}
    \label{fig:political-leaning-all}
\end{figure}

\begin{figure}
    \centering
    \includegraphics[width=\columnwidth]
    {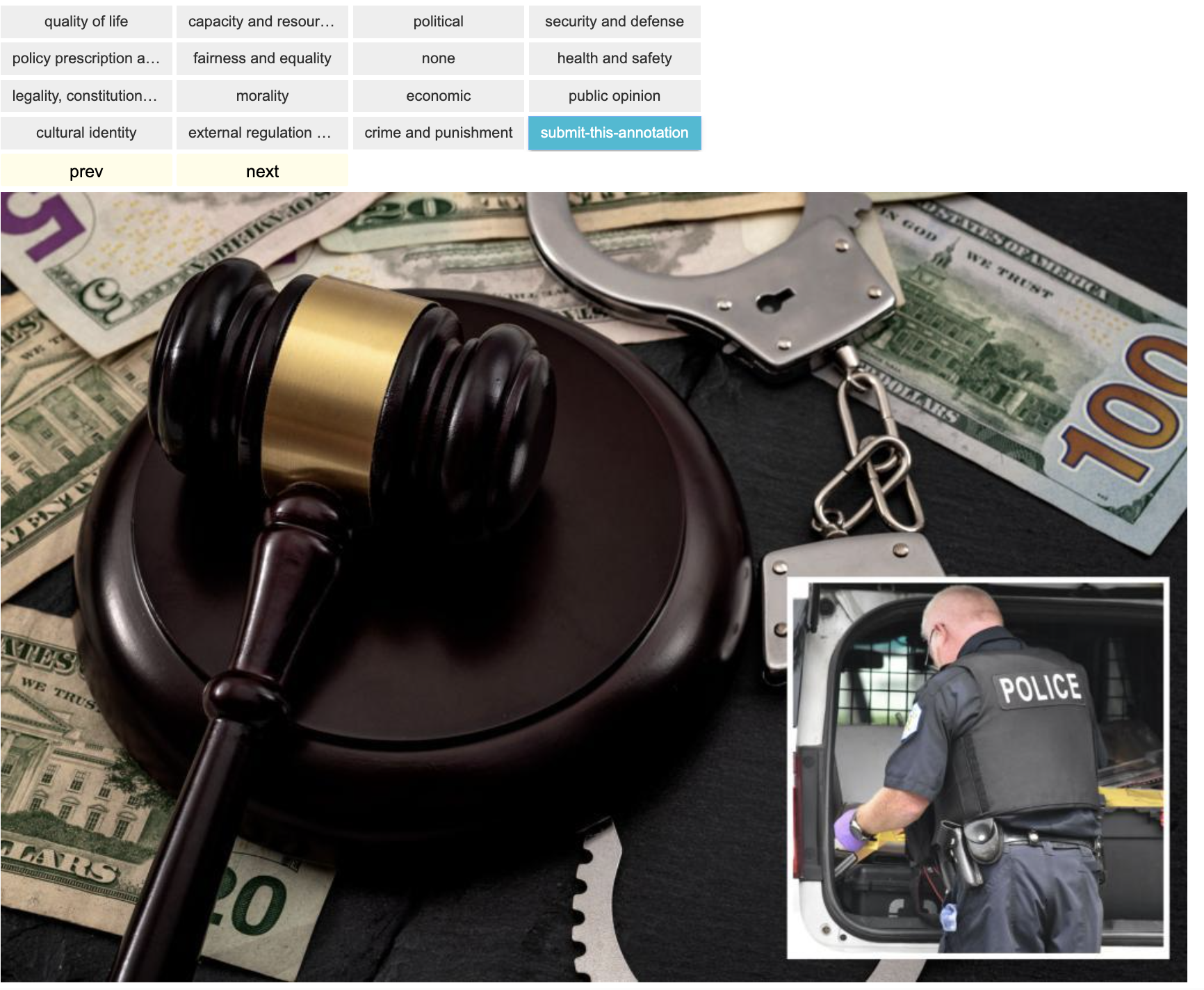}
    \caption{Image annotation UI used for annotating images frames by annotators}
    \label{fig:image-annotation-ui}
\end{figure}

\begin{figure}
    \centering
    \begin{subfigure}[t]{.45\linewidth}
        \includegraphics[width=\linewidth]{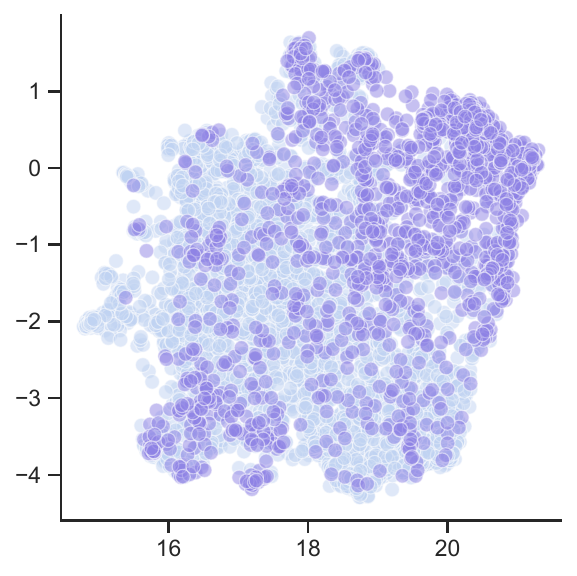}
        \caption{images}
    \end{subfigure}
    \begin{subfigure}[t]{.45\linewidth}
        \includegraphics[width=\linewidth]{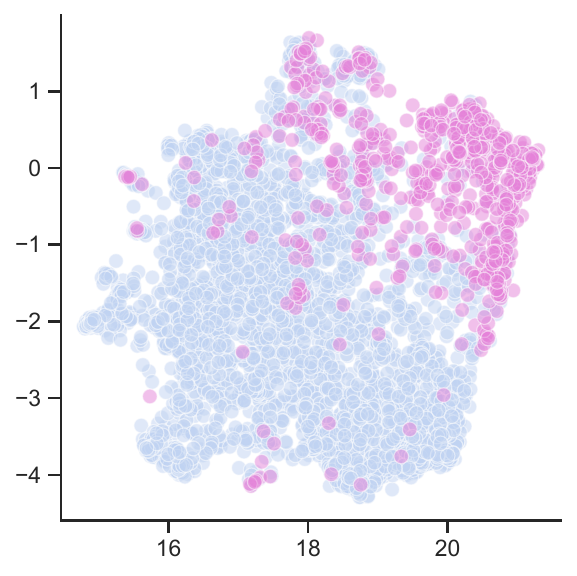}
        \caption{texts}
    \end{subfigure}
    \caption{Image and text frames cover different parts of the topic space. UMAP reduction (n\_neighbors=200, cosine distance) of a 5k sample of the generated topic descriptions (TF-IDF weighted token vectors: max\_features=5000, min\_df=5, max\_df=0.95) of the articles. Highlighted points represent the \textit{political} frame.}
    \label{fig:umap-plot-img-vs-txt}
\end{figure}

\lstdefinestyle{mystyle}{
    backgroundcolor=\color{backcolour},   
    commentstyle=\color{codegreen},
    keywordstyle=\color{magenta},
    numberstyle=\tiny\color{codegray},
    stringstyle=\color{codepurple},
    basicstyle=\ttfamily\footnotesize,
    breakatwhitespace=true,         
    breaklines=true,
    captionpos=b,                    
    keepspaces=true,                 
    numbers=none,                    
    numbersep=5pt,                  
    showspaces=false,                
    showstringspaces=false,
    showtabs=false,                  
    tabsize=2,
    xleftmargin=2em,
    xrightmargin=2em
}

\lstset{style=mystyle}
\begin{figure*}[t]
\begin{lstlisting}[label=lst:prompt-image-frame-prediction, caption=The prompt template for frame prediction using the pixtral model, numbers=none,basicstyle=\ttfamily\tiny]

SYS_PROMPT = f"""You are an intelligent and logical journalism scholar conducting analysis of images associated with news artciles."""

prompt_entity = """Your task is to identify the main subject or entity in the image that is central to the image. Entities are people or organizations that play a central role in the image. If there are multiple entities, choose the one that is central to the image. if there are no central or clearly identifiable entities, choose "None".
Additionally, analyse the image and output the sentiment with which the subject is portrayed in the image. The sentiment can be "positive", "negative" or "neutral". In case of no entity, output "None. Output only the json and no other text.
Format your output as a json entry as follows:
{"entity-name": "<name of the entity>", "sentiment": "<sentiment towards the entity>", "sentiment-reason": "<reasoning for the portrayed sentiment>"}
<image>
For the image above, provide the name of the entity, sentiment towards the entity, and reasoning for the chosen sentiment."""
FRAMES = f"""
Economic - costs, benefits, or other finance related. The image can includes things including but not limited to  money, funding, taxes, bank, meetings with a logo of a financial institution. If you are using logo of a financial instituion to classify it as economic, make sure it is clearly visible. If it is not clearly visible, it should be classified as 'None'. A professional attire in itself doesn not mean economic frame. 
Capacity and resources - availability of physical, human, or financial resources, and capacity of current systems. In the image, we can see things including but not limited to a geographical area, farmland, agriculture land, labour, people working in an institution, or images that convey scarcity or surplus in some way. 
Morality - religious or ethical implications. In the image, we can see things including but not limited to god, death, priests, church, protests related to moral issues.
Fairness and equality - balance or distribution of rights, responsibilities, and resources. In the image, we can see things including but not limited to the fight for civil or political rights, LGBTQ,  or calls to stopping discrimination.
Legality, constitutionality and jurisprudence - legal rights, freedoms, and authority of individuals, corporations, and government. In the image, we can see things including but not limited to , prisons, laws, judges in robes, courtrooms, legal documents, and prison facilities. This does not include sports contexts, such as referees or players enforcing or breaking game rules.
Policy prescription and evaluation - discussion of specific policies aimed at addressing problems. In the image, we can see things including but not limited to discussions on rule, rule making bodies, people in formal settings such as boardrooms or legislative halls - actively debating, and reviewing policy drafts or proposals. You might see official charts, graphs, or official documents. People in formal attire with no other information should not be classified as policy prescription and evaluation.
Crime and punishment - effectiveness and implications of laws and their enforcement. In the image, we can see things including but not limited to criminal activities, violence, police officers making arrests, crime scenes with investigators, courtrooms during criminal trials, prisons with detainees. This frame specifically excludes contexts involving sports, such as referees, players, or rule enforcement within games, which are not related to societal law violations or legal punishment.
Security and defense - threats to the individual, community, or nation. In the image, we can see things including but not limited to military uniforms, defense personnel, border patrol, war, soldiers, military equipment like tanks or fighter jets, border walls, or surveillance systems monitoring wide areas.
Health and safety - health care, sanitation, public safety. Images with objects like coffee, drinks, food items or activities like sports which a clear and literal message that it affects health and safety positively or negatively should be classified as health and safety, otherwise it should be classified as 'None'. E.g. a person drinking coffee does not mean health and safety, but a person drinking a medicine or having cigarette does. A bus does not mean health and safety, but a bus with a warning sign does.  In the image, we can see things including but not limited to doctors, nurses, injury, disease, or events with environmental impact that may impact health and safety. 
Quality of life - threats and opportunities for the individual's wealth, happiness, and well-being. In the image, we can see things that improves happiness or demonstrates quality of life in some form. It also includes things that demonstrate deterioration of quality of life by showing hardships of people, homelessness etc. This may also include happy children, food items that demonstrate good quality of life or people enjoying a nice meal.
Cultural identity - traditions, customs, or values of a social group in relation to a policy issue. In the image, we can see things including but not limited to concerts, cultural dance, sports, art, celebrities, artists and prominent people related to these topics. Examples, celebrities, traditional dress, sports with clear countriy specific detail e.g. jerseys/flags, culural events, cultural art etc. Otherwise, it should be classified as 'None'.
Public opinion - attitudes and opinions of the general public, including polling and demographics. Includes generic protests, people (non-celebrities) engaging with large crowds, riots, and strikes and including but not limited to sharing petitions and encouraging people to take political action. It will also include news broadcasts, talk shows, and interviews with people that are related to public opinion at large. 
Political - considerations related to politics and politicians, including lobbying, elections, and attempts to sway voters. In the image, we can see things related to politicians, elections, voting, political campaigns. Just formal clothing does not mean political frame. If the images does not have a political person which is recognizable, it should not be classified as political. A formal attire with no political information should be classified as 'None'.
External regulation \& reputation - international reputation or foreign policy. In the image, we can see things including but not limited to international organizations, global discussions/meetings, foreign policy, flags from multiple countries, or delegates at a cross-country forum discussing reputation and regulation. If you use a logo of a global organization to classify it as external regulation and reputation, make sure it is clearly visible in the image. If it is not clearly visible, it should be classified as 'None'.
None - no frame could be identified because of lack of information in the image. This should be selected when no other frame is applicable. Example, a handshake with no other information, a logo of a company with no other information, a landscape with no other information, a person in a photo album with no other information, a person speaking with no other information about the content of the speech or person's identity, a formal event with  no other information, a person in formal attire with no other information, a news logo with no news, a sports event with no additional information, simple objects like vehicle/car/pen/paper/sign-boards/objects etc with no other information etc."""

FRAMING_PROMPT = "A set of generic news frames with an id, name and description are: \n"
FRAMES_TASK_PROMPT = """
Given the list of frames, and the image.
<image>
Your task is to carefully analyse the image and choose the appropriate frames from the above list.
Output your answer in a json format with the format:
{"frames-list": "[<All frame names that apply from list provided above>], "reason": "<reasoning for the frames chosen>"}
Output only the json and no other text.
"""
\end{lstlisting}
\end{figure*}

\lstset{style=mystyle}
\begin{figure*}[t]
\begin{lstlisting}[label=lst:text-frame-prediction, caption=The prompt template for text prediction using the mistral model, numbers=none,basicstyle=\ttfamily\tiny]

SYS_PROMPT = f"You are an intelligent and logical journalism scholar conducting analysis of news articles. Your task is to read the article and answer the following question about the article. Only output the json and no other text.\n"

TOPIC_PROMPT = "Output the topic of the article, along with a justification for the answer. The topic should be a single word or phrase. Format your output as a json entry with the field 'topic_justification' and 'topic'."

ENTITY_PROMPT = """Your task is to identify the main subject or entity in the article that is central to the article. Entities are people or organizations that play a central role. If there are multiple entities being discussed, choose the one that is central to the article. If there are no central or clearly identifiable entities, choose "None".
Additionally, analyse the image and output the sentiment with which the subject is portrayed in the image. The sentiment can be "positive", "negative" or "neutral". In case of no entity, output "None. Output only the json and no other text.
Format your output as a json entry as follows:

{"entity-name": "<name of the entity>", "sentiment": "<sentiment towards the entity>", "sentiment-reason": "<reasoning for the portrayed sentiment>"}

For the given article, provide the name of the entity, sentiment towards the entity, and reasoning for the chosen sentiment."""

FRAMES = """
A list of frame names and their descriptions used in news is:
Economic - costs, benefits, or other financial implications,
Capacity and resources - availability of physical, human, or financial resources, and capacity of current systems, 
Morality - religious or ethical implications,
Fairness and equality - balance or distribution of rights, responsibilities, and resources,
Legality, constitutionality and jurispudence - rights, freedoms, and authority of individuals, corporations, and government,
Policy prescription and evaluation - discussion of specific policies aimed at addressing problems,
Crime and punishment - effectiveness and implications of laws and their enforcement,
Security and defense - threats to welfare of the individual, community, or nation,
Health and safety - health care, sanitation, public safety,
Quality of life - threats and opportunities for the individual's wealth, happiness, and well-being,
Cultural identity - traditions, customs, or values of a social group in relation to a policy issue,
Public Opinion - attitudes and opinions of the general public, including polling and demographics,
Political - considerations related to politics and politicians, including lobbying, elections, and attempts to sway voters,
External regulation and reputation - international reputation or foreign policy of the U.S,
None - none of the above or any frame not covered by the above categories."""
    
GENERIC_FRAMING_PROMPT = f"""Framing is a way of classifying and categorizing information that allows audiences to make sense of and give meaning to the world around them (Goffman, 1974).
Entman (1993) has defined framing as "making some aspects of reality more salient in a text in order to promote a particular problem definition, causal interpretation, moral evaluation, and/or treatment recommendation for the item described".
Frames serve as metacommunicative structures that use reasoning devices such as metaphors, lexical choices, images, symbols, and actors to evoke a latent message for media users (Gamson, 1995).
A set of generic news frames with an id, name and description are: {FRAMES}.
Your task is to code articles for one of the above frames and provide justification for it. Format your output as a json entry with the fields 'frame_justification', 'frame_id', 'frame_name'.
'frame_name' should be one of the above listed frames. Only output one frame per article."""

GENERIC_FRAMING_MULTIPLE_PROMPT = """
Given the list of news frames, and the news article.
Your task is to carefully analyse the article and choose the appropriate frames used in the article from the above list.
Output your answer in a json format with the format:
{"frames-list": "[<All frame names that apply from list provided above>], "reason": "<reasoning for the frames chosen>"}.
Only choose the frames from the provided list of frames. If none of the frames apply, output "None" as the answer.
"""

ISSUE_FRAMING_PROMPT = """
Entman (1993) has defined framing as "making some aspects of reality more salient in a text in order to promote a particular problem definition, causal interpretation, moral evaluation, and/or treatment recommendation for the item described".
Frames serve as metacommunicative structures that use reasoning devices such as metaphors, lexical choices, images, symbols, and actors to evoke a latent message for media users (Gamson, 1995).
There are several ways to cover a specific issue in the news. For instance, the issue of climate change can be framed as a scientific, a political, a moral, or a health issue etc. with issue-specific frames such as "Global Doom", "Local Tragedies", "Sustainable future".
Similarly, articles related to immigration can frame immigrants as a hero, a victim, or a threat with frames such as "Economic Burden", "Cultural Invasion", "Humanitarian Crisis".
Based on the topic of the article, come up with an issue-specific frame that is relevant to the topic of the article. Provide a justification for the frame. 
Format your output as a json entry with the fields 'issue_frame_justification' and 'issue_frame'."""

POST_PROMPT = " Output only the json and no other text. Make sure to add escape characters where necessary to make it a valid json output."
\end{lstlisting}
\end{figure*}

\begin{table*}[h]
    \centering
    \footnotesize
    \begin{tabular}{llp{11cm}}
        \toprule
        \textbf{Topic Name} & \textbf{Accuracies} & \textbf{Example} \\
        \midrule
        \vspace{0.5em}
        \textbf{Accident} & 90\%, 100\% & \textit{YONKERS, N.Y. -- A 70-year-old woman was struck and killed by a car while walking on the sidewalk in Yonkers. It happened Sunday night on North Broadway. Investigators say the car then went over a retaining wall...} \\ \vspace{0.5em}
        
        \textbf{Crime} & 90\%, 90\% & \textit{MINNEAPOLIS -- The person suspected of causing a crash that killed five young women is in custody at the Hennepin County Jail. WCCO is not naming the man until he's charged with a crime, which prosecutors say could happen as soon as Tuesday...} \\ \vspace{0.5em}

        \textbf{Culture} & 70\%, 70\% & \textit{CHICAGO (CBS) -- You can start your summer with a pop of color at the new Andy Warhol exhibition on the campus of the College of DuPage. With more than 200 original photographs...} \\ \vspace{0.5em}

        \textbf{Economy} & 100\%, 100\% & \textit{Andrew Ross Sorkin grilled White House economic advisor Heather Boushey on Wednesday over whether the Biden administration planned for increased inflation when the president passed several spending packages. The panel was discussing....} \\ \vspace{0.5em}

        \textbf{Education} & 100\%, 90\% & \textit{BALTIMORE - Westminster National Golf Course hosted 100 third graders from Westminster Elementary School on Thursday for a hands-on cross-curricular STEM-related field trip. The students learned all about golf and the science...} \\ \vspace{0.5em}

        \textbf{Entertainment} & 100\%, 100\% & \textit{A new animated fantasy comedy movie that follows the adventures of a preteen Latina who wants to do her own thing while surrounded by her multigenerational Mexican American family premieres Friday on Netflix...} \\ \vspace{0.5em}

        \textbf{Finance} & 100\%, 100\% & \textit{WASHINGTON, May 30 (Reuters) - The former head of Wells Fargo \& Co's (WFC.N) retail bank agreed to pay a \$3 million penalty to settle the U.S. Securities and Exchange Commission's fraud charges for misleading investors...} \\ \vspace{0.5em}

        \textbf{Health} & 100\%, 90\% & \textit{LOS ANGELES (AP)‚ Madonna has postponed her career-spanning Celebration tour due to what her manager called a ``serious bacterial infection.'' Manager Guy Oseary wrote on Instagram Wednesday that the singer had spent several days... } \\ \vspace{0.5em}

        \textbf{Immigration} & 100\%, 100\% & \textit{A U.S. Citizenship and Immigration Services (USCIS) district office in New York City. USCIS expects to accept and approve a low number of H-1B registrations from the H-1B lottery‚ first selection round...} \\ \vspace{0.5em}

        \textbf{Legal} & 80\%, 80\% & \textit{Former President Donald Trump stubbornly rejected his legal team's efforts last year to settle the classified documents case and prevent him from being indicted by a federal grand jury, according to a bombshell report. Christopher Kise, one of Trump's attorneys in the fall of 2022...} \\ \vspace{0.5em}

        \textbf{Lifestyle} & 80\%, 80\% & \textit{The courgettes are roasting sweetly in the oven, half of them for lunch today dressed with sultanas, pine kernels and honey, the rest to serve as a salad tomorrow. This is something I also do with aubergines, red onions and sweet potatoes. There are so many vegetables to roast right now...} \\ \vspace{0.5em}

        \textbf{Politics} & 90\%, 100\% & \textit{Rep. Matt Gaetz (R-Fla.) criticized House Republicans' recent effort to impeach President Joe Biden saying it was not done in a ``legitimate'' or ``serious'' way, a video obtained by NBC News shows, raising questions of whether he will support Rep. Jim Jordan (R-Ohio), who is a leading candidate to become House speaker...} \\ \vspace{0.5em}

        \textbf{Safety} & 90\%, 50\% & \textit{The Sacramento region has some of the highest numbers of fatal traffic collisions in the state. Sacramento police say that last year, more than 50 people died on city streets. Now, as part of National Passenger Safety Week, there's an effort to reduce fatal collisions...} \\ \vspace{0.5em}

        \textbf{Social Issues} & 70\%, 70\% & \textit{REDWOOD CITY, The San Mateo County Board of Supervisors is continuing to explore ways to provide more housing for farmworkers in the county, nearly four months after a mass shooting in Half Moon Bay exposed an urgent need for more living options for agricultural workers with low income...} \\ \vspace{0.5em}

        \textbf{Technology} & 70\%, 100\% & \textit{America's spending on artifical intelligence in public safety is projected to increase from $9.3 billion in 2022 to $71 billion by 2030, according to a new analysis by the Insight Partners research firm. The projected seven-year boom is expected to be fueled by global and domestic terrorism...} \\ \vspace{0.5em}

        \textbf{War} & 100\%, 100\% & \textit{The Ukrainian air-assault force‚Äôs 25th Brigade just got a lot heavier. Photos that appeared online this week confirm the brigade has re-equipped with German-made Marder infantry fighting vehicles. The 25th is the second Ukrainian air-assault brigade, after the 82nd, to get 31-ton Marders from German stocks...} \\ \vspace{0.5em}

        \textbf{Weather} & 100\%, 100\% & \textit{ Tropical Storm Nigel is expected to become a hurricane as soon as Monday, the National Hurricane Center said Sunday, and could be the latest tropical storm in the Atlantic this season to rapidly intensify to major hurricane status...} \\ 
        
        \bottomrule
    \end{tabular}
    \caption{The 19 topics used in our analysis. Accuracies represent the average acceptability judgments of two annotators (two of the paper authors) over a set of 10 predicted examples for each topic.}
    \label{table:topics}
\end{table*}

\end{document}